\documentclass[preprint,10.5pt]{elsarticle}
\usepackage[margin=1in]{geometry}
\pdfoutput=1 

\usepackage{amssymb,amsmath,amsthm,amsbsy}
\usepackage{physics}
\usepackage[pdftex,bookmarksnumbered]{hyperref}

\usepackage{amssymb,amsmath,amsthm,amsbsy}
\usepackage{systeme}
\usepackage{physics}
\usepackage{hyperref} 
\usepackage{graphicx}
\usepackage{epstopdf}
\usepackage{epsfig}
\usepackage{multicol}
\usepackage{subfigure}
\usepackage{enumitem}
\usepackage{multirow}
\usepackage{algorithm,algorithmic}
\usepackage{bm}
\usepackage{setspace}
\usepackage{color}
\usepackage[mathlines,running]{lineno}

\DeclareMathOperator*{\argmin}{arg\,min}

\newcommand{\paolo}[1]{\textcolor{black}{#1}}

\newcommand{\cA}{{\cal A}}

\newcommand{\micr}{\,$\mu$m\,}

\newcommand{\bfm}[1]{\mathbf{#1}}
\newcommand{\bsn}[1]{\boldsymbol{#1}}

\newcommand{\bfzero}{\bfm{0}}

\newcommand{\bfB}{\bfm{B}}

\newcommand{\bfu}{\bfm{u}}

\newcommand{\bfS}{\bfm{S}}
\newcommand{\bfX}{\bfm{X}}

\newcommand{\bfv}{\bfm{v}}
\newcommand{\bfP}{\bfm{P}}

\newcommand{\bfF}{\bfm{F}}
\newcommand{\bfG}{\bfm{G}}

\newcommand{\bfM}{\bfm{M}}
\newcommand{\bfN}{\bfm{N}}
\newcommand{\bfK}{\bfm{K}}

\newcommand{\bfC}{\bfm{C}}
\newcommand{\bfH}{\bfm{H}}

\newcommand{\bfE}{\bfm{E}}

\newcommand{\bfphi}{\bsn{\phi}}

\newcommand{\bfbeta}{\bsn{\beta}}

\newcommand{\bfmu}{\bsn{\mu}}

\journal{Computer Methods in Applied Mechanics and Engineering}

\begin{document}
\makeatletter
\def\ps@pprintTitle{%
  \let\@oddhead\@empty
  \let\@evenhead\@empty
  \let\@oddfoot\@empty
  \let\@evenfoot\@oddfoot
}
\makeatother
\begin{frontmatter}
\title{ 
Reduced order modeling of parametrized systems through autoencoders and SINDy approach:
continuation of periodic solutions
}


\author[dica]{Paolo Conti}\corref{cor1} 
\author[dica]{Giorgio Gobat}
\author[mox]{Stefania Fresca}
\author[mox]{Andrea Manzoni}
\author[dica]{Attilio Frangi}

\address[dica]{Department of Civil Engineering, Politecnico di Milano}
\address[mox]{MOX -- Department of Mathematics, Politecnico di Milano}

\begin{abstract}
Highly accurate simulations of complex phenomena governed by partial differential equations (PDEs) typically require intrusive methods and entail expensive computational costs, which might become prohibitive when approximating steady-state solutions of PDEs for multiple combinations of control parameters and initial conditions. Therefore, constructing efficient reduced order models (ROMs) that enable accurate but fast predictions, while retaining the dynamical characteristics of the physical phenomenon as parameters vary, is of paramount importance. In this work, a data-driven, non-intrusive framework which combines ROM construction with reduced dynamics identification, is presented. \paolo{Starting from a limited amount of full order solutions, the proposed approach leverages autoencoder neural networks with parametric sparse identification of nonlinear dynamics (SINDy) to construct a low-dimensional dynamical model. This model can be queried to efficiently compute full-time solutions at new parameter instances, as well as directly fed to continuation algorithms.} 
These aim at tracking the evolution of periodic steady-state responses as functions of system parameters, avoiding the computation of the transient phase, and allowing to detect instabilities and bifurcations. Featuring an explicit and parametrized modeling of the reduced dynamics, the proposed data-driven framework presents remarkable capabilities to generalize with respect to both time and parameters. Applications to structural mechanics and fluid dynamics problems illustrate the effectiveness and accuracy of the proposed method.

\end{abstract}




\end{frontmatter}





\section{Introduction}
\label{sec:introduction}

\noindent The ever-increasing resources and computational power allow nowadays to simulate very complex physical phenomena  more and more accurately, even in multi-scale and multi-physics scenarios. However, the solution of parametrized, time-dependent systems of partial differential equations (PDEs) by means of full order models (FOMs) -- such as the finite element method -- may clash with time and computational budget restrictions. Moreover, using FOMs to explore different scenarios with varying initial conditions and parameter combinations might be a computationally prohibitive task, or even infeasible in several 
practical applications. 
\paolo{Differently from the problem of estimating output quantities of interest that depend on the solution of the differential problem, the computation of the whole solution field is intrinsically high-dimensional, with additional difficulties related to the nonlinear and time-dependent nature of the problem.} 
%
All these reasons drive the search of efficient, but accurate, reduced order models (ROMs). Among these, the reduced basis method \cite{quarteroni2015reduced,HesthavenRozzaStamm,benner2017model} is a very well-known approach,  exploiting, e.g., proper orthogonal decomposition (POD) to build a reduced space, either global or local \cite{amsallem2012nonlinear,pagani2018numerical}, to approximate the solution of the problem. 
However, despite their accuracy and mathematical, these techniques are in general intrusive \cite{gobat2022reduced}. 

Among machine and deep learning techniques widely used to build surrogate models or emulators to the solution of parametrized, nonlinear, time-dependent system of PDEs, autoencoder (AE) neural networks \cite{goodfellow2016deep} have recently become a popular strategy because they allow to non-intrusively reduce dimensionality and unveil latent features directly from data streams, without accessing the FOM operators \cite{gonzalez2018deep,lee2020model,maulik2021reduced,kim2022fast}. 
Their success is due to the expressiveness capacity of neural networks \cite{cybenko1989approximation, leshno1993multilayer, hornik1989multilayer}, which enables outstanding performances in nonlinear compression and great flexibility in identifying coordinate transformations \cite{lusch2018deep}.
\paolo{Recently proposed techniques consider convolutional AEs, suitably coupled to deep feedforward neural networks in order to simultaneously perform dimensionality reduction and learn the parameter-to-solution map, as in the case of deep learning-based ROMs (DL-ROMs) \cite{fresca2021comprehensive,franco2021deep}, and their enhanced version through POD (POD-DL-ROMs) \cite{fresca2022pod}. These techniques outperform (in some cases, also by several  orders of magnitude) classical ROMs in terms of both the dimension of the reduced problem and the query time for predicting the solution at unseen instances.}
%
The resulting ROMs are nonintrusive and low-dimensional,  ultimately enabling real-time simulations in complex physical scenarios; 
%
%
%
see, e.g., \cite{fresca2022deep} for a possible application to the real-time simulation of the mechanical behavior of micro-electro-mechanical systems (MEMS). We highlight that this approach relies on the idea that the system dynamics lies on a low dimensional invariant manifold, setting a clear parallel with the Direct Parametrization of Invariant Manifolds approach preconized, e.g., in \cite{vizzaccaro2021direct,opreni2021model,vizzaccaro2022high}; see also \cite{gobat2022virtual} for further details.  

Nonetheless, this approach relies on the underlying physical model only when computing FOM snapshots required for training the deep neural networks involved, but not when querying the parameter-to-solution map for unseen parameter values at testing stage. Moreover, although showing impressive efficiency during both the training and the testing phases, and providing reliable predictions on unseen scenarios in the time-parameter range where training data are generated, POD-DL-ROMs \paolo{show some limitations in terms of interpretability} and generalization capability outside the observed parametric-temporal domain.   

A possible alternative approach to learn the reduced space dynamics is represented by the so-called Sparse Identification of Dynamical Systems (SINDy) method, a sparse regression technique initially introduced for learning dynamics from time-series data, and then further generalized to address several features  \cite{kaiser2018sparse,brunton2016discovering, brunton2016sparse, goyal2022discovery}. In the past few years, SINDy has been widely applied to identify models of fluid flows, convection phenomena, structural models, and many others.
The appeal of the SINDy approach is the generation, from a dictionary of pre-defined (analytical) functions, an explicit ROM in the form of a system of ordinary differential equations (ODEs), that can improve the interpretability of the latent space dynamics thanks to suitable sparsity constraints. Regarding time integration, as well as the computation of derived output quantities of interest, the identified dynamical system can be then treated with standard numerical tools. For instance, when the transient dynamics is relevant, or when steady-state regimes are not required, the obvious choice is represented by time-marching techniques. However, this might not be the optimal choice when periodic solutions are needed. These latter arise in many practical applications,  ranging from aircraft design and simulation \cite{ananthkrishnan1996characterization}, chemistry \cite{epstein1996nonlinear} to MEMS  \cite{opreni2021analysis}, just to mention some instances. Periodicity conditions can be enforced directly in the formulation of the dynamical system and ad-hoc numerical methods can be applied like, e.g., the Harmonic Balance \cite{krack2019harmonic} or collocation techniques \cite{doedel2007auto}. While their computational burden is usually prohibitive for large-scale FOMs \cite{detroux2015harmonic,opreni2021analysis}, these techniques become much more appealing when used in conjunction with ROM strategies.

Even if the SINDy approach has been so far mainly applied to rather low-dimensional problems, Champion et al. in \cite{champion2019data} have fostered its use in combination with AE neural networks, laying the foundation for a unified framework for dimensionality reduction and system identification. This approach with recent extensions \cite{bakarji2022discovering,goyal2021learning, kneifl2021nonintrusive, callaham2022role, fukami2021sparse}, however, does not encompass the possibility to treat parameters and forcing dependencies. This inevitably precludes the possibility of drawing efficiently a complete portrait of the dynamics of the observed system in multiple scenarios. In this respect, the SINDy method offers a great  flexibility in view of incorporating parameters and forcing terms in the identified ODE system \cite{brunton2016sparse}. The major benefit of identifying an explicit parameter-dependency in the latent system is represented by the possibility to use \textit{continuation} algorithms to track the evolution of periodic responses as a function of system parameters 
avoiding the computation of transients, which represents a significant computational advantage for long transient systems or for real-time applications. Moreover, these algorithms allow to identify, in a straightforward way, instabilities and bifurcations which might occur in the latent dynamics and result in macroscopic changes of the steady-state behavior of the physical system. \paolo{For example, variations of input parameters can result in the development of vorticity in fluids or resonance phenomena in mechanical structures. In practical applications, these behaviors can significantly affect the operation of the observed systems, therefore it is essential to estimate at which parameter configurations these regime transitions occur.} 

In the present work, we propose an extension of the AE+SINDy approach presented in \cite{champion2019data} to efficiently construct reduced order approximations of parametrized PDE solutions, as well as to discover their underlying dynamics and parametric dependencies, starting from a limited number of FOM snapshots. Our method hinges upon a data-driven, non-intrusive, ROM construction employing POD and AEs -- the one provided by the so-called POD-DL-ROMs \cite{fresca2022pod} -- together with reduced dynamics identification through a parametrized SINDy approach. The proposed framework thus leverages recent SINDy extensions with the aforementioned POD-DL-ROM reduction. In contrast to recent works in the same direction \cite{fries2022lasdi,kalia2021learning}, the main appeal of our method is that, by representing the latent dynamics as a parametrized system of ODEs, we can naturally embed continuation algorithms in the overall framework, thus providing a comprehensive description of the dynamics of the physical system.

The proposed method is first applied to a structural mechanics problem dealing with a straight beam MEMS resonator, which is excited at different forcing amplitudes and frequencies. 
We show how the proposed strategy allows to construct a ROM which manages both to generalize with respect to parameters and to accurately forecast the solution over long term horizons, even reaching the \textit{steady-state} response for which no measurement is provided. In addition, the nonlinear hardening behaviour of the beam is portrayed by means of continuation algorithms directly from the reduced identified system.
Next, the problem of a fluid flow past a cylinder is considered, employing the proposed method to efficiently approximate the fluid velocity and pressure at different regimes as the Reynolds number varies, as well as to empirically identify, through continuation algorithms, the bifurcation value of the transition from laminar to unsteady behaviour.

\paolo{Besides the considered examples, the AE+SINDy method can be applied straightforwardly to any time series which tends to a periodic regime. Moreover, although this work focuses on periodic orbits, the entire framework could be also extended to handle quasi-periodic solutions by employing continuation techniques which are suitable for these regimes \cite{guillot2017continuation}, or, more in general, to any time series that tends towards a limit cycle -- which is the only requirement for continuation algorithms to work \cite{krauskopf2007numerical}. 
}

The paper is structured as follows. In Sect. \ref{sec: method} we detail the structure of the method and we present how to use it for the efficient approximation of parametrized PDE solutions as well as for dynamic analysis using both time-marching schemes and continuation algorithms. In Sect. \ref{sec:applications} we present and discuss its performance on the two aforementioned numerical examples, finally drawing some concluding remarks in Sect. \ref{sec:conclusions}.

\section{Reduced order modeling with AE and SINDy}
\label{sec: method}

\subsection{Problem setup}
\noindent Let us consider the following general form of a dynamical system
\begin{equation}
    \begin{cases} 
    \dot{\vb{x}}(t; \bm{\beta}) = \vb{f}(t,\vb{x}(t; \bm{\beta});\bm{\beta}), & t \in (0,T),\\ 
    \vb{x}(0;\bm{\beta}) = \vb{x}_0,
    \end{cases}
\label{eq: dy_eq_full}
\end{equation}
where $\vb{x}\in \mathbb{R}^{N}$ is the state of the system, $\dot{\vb{x}}$ its time derivative, $t$ the time, $\vb{x}_0$ the initial state, while $\vb{f}$ is the function that defines the dynamical evolution of the physical system. Here, $\bm{\beta}\in \mathbb{R}^{p}$ represents the vector collecting $p$ (possibly) time dependent parameters and/or forcing terms. 
System \eqref{eq: dy_eq_full} consists of a set of ordinary differential equations (ODEs), whose dimensionality $N$ represents the number of degrees of freedom associated to the space discretization technique (such as, for instance, finite element, finite volume or isogeometric analysis methods) applied to a system of partial differential equations (PDEs) which defines the physical system.
Therefore, the size $N$ is typically extremely large and the system \eqref{eq: dy_eq_full} is denoted as full order model (FOM). 

The goal is to learn the dynamics $\vb{f}$ from a set of time histories of the state $\vb{x}$ and its time-derivative $\dot{\vb{x}}$ for different instances of $\bm{\beta}.$ In general,  $\dot{\vb{x}}$ can be either computed directly or approximated numerically from $\vb{x}$. Snapshots of the time histories are stacked in matrices 
\[
\mathbf{X}=\left[\begin{array}{c|c|c|c|c|c|c|c|c}
\vb{x}(t_1;\bm{\beta}_1) & \vb{x}(t_2; \bm{\beta}_1)  & \ldots & \vb{x}(t_{N_t};\bm{\beta}_1)  & \vb{x}(t_{1};\bm{\beta}_2)  & \ldots &  \vb{x}(t_{N_t};\bm{\beta}_2) & \ldots & \vb{x}(t_{N_t};\bm{\beta}_{N_\beta})
\end{array}\right]^T, 
\] 
\[
\dot{\mathbf{X}}=\left[\begin{array}{c|c|c|c|c|c|c|c|c}
\dot{\vb{x}}(t_1;\bm{\beta}_1) & \dot{\vb{x}}(t_2; \bm{\beta}_1)  & \ldots & \dot{\vb{x}}(t_{N_t};\bm{\beta}_1)  & \dot{\vb{x}}(t_{1};\bm{\beta}_2)  & \ldots &  \dot{\vb{x}}(t_{N_t};\bm{\beta}_2) & \ldots & \dot{\vb{x}}(t_{N_t};\bm{\beta}_{N_\beta})\\
\end{array}\right]^T, 
\]
\noindent respectively, such that $(\vb{X},\dot{\vb{X}}) \in \mathbb{R}^{N_t N_\beta \times N}$, where $N_t$ is the total number of time instants and $N_\beta$ is the number of parameter/forcing instances considered. However, approximating such a high dimensional function $\vb{f}$ might be an extremely complicated and computationally demanding task.




\subsection{Latent space construction}
\noindent In order to significantly speed up the computationally expensive simulations, to have a more interpretable and parsimonious dynamic model, and to efficiently estimate the dynamical behaviour of the system at steady-state by continuation,
we propose a reduction technique that shall  simultaneously reduce 
the problem dimensionality and provide a new set of coordinates better suited to represent the dynamics of the problem. 
Starting from the assumption that the full order solution of the parametrized system \eqref{eq: dy_eq_full} lies on a low-dimensional manifold embedded in high-dimensional (discrete) space \cite{opreni2021model}, we efficiently reduce the dimensionality of the problem directly from the snapshot data matrix $\mathbf{X}$, without accessing the FOM operators appearing in \eqref{eq: dy_eq_full}.
The reduced space dimensionality $n$ can be set equal to the intrinsic dimension of the solution manifold whenever this is known a priori or, conversely, it can be tuned to discover the intrinsic dimension of an unknown system for which only state measurements are available.

Therefore, we consider a preliminary linear reduction by POD, projecting the full order states of the system $\vb{x}$ onto the $N_\text{POD}$ dominant \textit{singular vectors} denoted as $\tilde{\vb{U}}\in \mathbb{R}^{N\times N_\text{POD}}$, i.e., $\tilde{\vb{x}}=\vb{x}\tilde{\vb{U}}$\footnotemark{}. POD coordinates of FOM data matrices are then $\tilde{\vb{X}}=\vb{X}\tilde{\vb{U}}, \dot{\tilde{\vb{X}}}=\dot{\vb{X}}\tilde{\vb{U}}$.
\footnotetext{In the present work, snapshots are stacked as rows for consistency with how neural networks are trained, and in contrast with standard POD literature, where snapshots are listed in columns. For clarification, the reader must intend POD to be applied to $\mathbf{X}^T$, i.e.\ $\mathbf{X}^T \approx \tilde{\mathbf{U}}\tilde{\mathbf{\Sigma}}\tilde{\mathbf{V}}^T$. Consistently with the notation adopted, POD projections -- usually indicated as $\tilde{\vb{x}} = \tilde{\mathbf{U}}^T\vb{x}$ -- are rewritten here in terms of row vectors, namely $\tilde{\vb{x}}=\vb{x}\tilde{\vb{U}}$.}
Next, a further nonlinear reduction is performed by means of an autoencoder neural network. In particular, we identify a new set of latent variables $\vb{z}$ such that 
\begin{equation}
    \vb{z}(t;\bm{\beta}) = \bm{\varphi}(\tilde{\vb{x}}(t;\bm{\beta})),
    \label{eq: reduction}
\end{equation}
where $\bm{\varphi}(\cdot) = \bm{\varphi}(\cdot\,;  \vb{W}_\varphi):\mathbb{R}^{N_\text{POD}}\rightarrow \mathbb{R}^{n}$, $n \ll N_\text{POD} \ll N$, is an \textit{encoder} consisting of a fully connected neural network and $\vb{W}_\varphi$ denotes the network parameters. 
Moreover, in order to go back to POD coordinates from intrinsic ones, we make use of a \textit{decoder}, that is, another fully connected neural network $\bm{\psi}(\cdot\,; \vb{W}_\psi):\mathbb{R}^n \rightarrow \mathbb{R}^{N_\text{POD}}$ such that
\begin{equation}
    \bm{\psi}(\vb{z}(t;\bm{\beta})) = \bm{\psi}(\bm{\varphi}(\tilde{\vb{x}}(t;\bm{\beta})))\approx \tilde{\vb{x}}(t;\bm{\beta}).
    \label{eq: autoencoder}
\end{equation}
Finally, to reconstruct the FOM solution, we project back through POD modes the output of the decoder, namely $\bm{\psi}(\vb{z}(t;\bm{\beta}))\tilde{\vb{U}}^T\approx \vb{x}(t;\bm{\beta})$.

The pair $\left(\bm{\varphi}, \bm{\psi}\right)$ constitutes the \textit{autoencoder} (AE) neural network \cite{goodfellow2016deep} which is trained at once by solving 
\begin{equation}
    \underset{\mathbf{W}_\varphi, \mathbf{W}_\psi}{\min  }\;\norm{   \tilde{\mathbf{X}} - \bm{\psi}\left(\bm{\varphi}\left(\tilde{\mathbf{X}};\mathbf{W}_\varphi\right);\mathbf{W}_\psi\right)}^2_{F},
    \label{eq: autoencoder_loss}
\end{equation}
where $\norm{\cdot}_F$ is the Frobenius norm and $\psi, \varphi$ are intended to be applied row-wise to the input matrices. The preliminary POD reduction enables a more efficient encoding process \cite{fresca2022pod}, since the AE only needs to reduce the dimension from a moderate number $N_\text{POD}$ of features to $n$,  instead of starting from an extremely large amount $N$ of degrees of freedom. 
It is worth stressing that POD, a linear reduction technique, presents several limitations when applied alone to nonlinear, time-dependent parametrized problems, thus motivating the 
inclusion of the neural network encoder. 

In the new set of intrinsic coordinates, defined by \eqref{eq: reduction}, system \eqref{eq: dy_eq_full} can be reformulated as
\begin{equation}
    \begin{cases} 
    \dot{\vb{z}}(t; \bm{\beta}) = \tilde{\vb{f}}(t,\vb{z}(t; \bm{\beta});\bm{\beta}), & t \in (0,T),\\ 
    \vb{z}(0;\bm{\beta}) = \vb{z}_0,
    \end{cases}
\label{eq: dy_eq_reduced}
\end{equation}
where $\tilde{\vb{f}}$ encodes the dynamics of the low-dimensional system, $\dot{\vb{z}} = \dot{\vb{\varphi}}(\tilde{\vb{x}}) = \dot{\tilde{\vb{x}}}\nabla^T_{\tilde{\vb{x}}}\bm{\varphi}(\tilde{\vb{x}})$ and $\vb{z}_0 = \bm{\varphi}(\tilde{\vb{x}}_0)$, with $\tilde{\vb{x}}_0=\vb{x}_0\tilde{\vb{U}}$ .

\subsection{Latent dynamics identification}
\noindent The problem of estimating the high-dimension function $\vb{f}\in \mathbb{R}^N$ in \eqref{eq: dy_eq_full} is now reduced to the estimation of $\tilde{\vb{f}}\in \mathbb{R}^n$ in \eqref{eq: dy_eq_reduced}, which describes the latent dynamics of the system in a low-dimensional space. 
We are interested in a model representation that should not only be accurate, but also interpretable and generalizable, that is, capable of forecasting in time and extrapolating with respect to parameters. For the latter reasons, a purely black-box approach is not optimal in this context. 

Therefore, to identify the latent dynamics, we make use of the SINDy regression strategy \cite{kaiser2018sparse,brunton2016discovering, brunton2016sparse},
which assumes that $\tilde{\vb{f}}$ can be expressed as a sparse combination of a set of linear and nonlinear candidate basis functions of the latent state $\mathbf{z}$ and of $\bm{\beta}$.
The choice of the library of candidate features is typically guided by a possible prior knowledge of the physical system and of the parameter/forcing dependency. A typical choice is polynomials as they frequently appear in most dynamical system models.

In practice, SINDy aims to approximate the latent dynamics $\tilde{\vb{f}}$ in \eqref{eq: dy_eq_reduced} as
\begin{equation}
    \dot{\vb{z}} = \tilde{\vb{f}}(\vb{z}; \bm{\beta})\approx \vb{\Theta}(\vb{z}, \bm{\beta})\mathbf{\Xi}
    \label{eq: SINDy}
\end{equation}
where $\vb{\Theta}(\vb{z},\bm{\beta})=\left[\,{\theta}_1(\vb{z},\bm{\beta}) \,,\,\ldots \,,\,{\theta}_r(\vb{z},\bm{\beta})\,\right]\in\mathbb{R}^r$ is the library of $r$ candidate functions to describe the dynamics of the data, while $\vb{\Xi}\in\mathbb{R}^{r\times n}$ is the unknown matrix of coefficients that determine the active terms from $\mathbf{\Theta}$ in the dynamics $\tilde{\vb{f}}$. Matrix $\mathbf{\Xi}$ is estimated by sparse regression
\begin{equation}
     \underset{\mathbf{\Xi}}{\argmin}\;\norm{{\mathbf{\dot{Z}} - \mathbf{\Theta}(\mathbf{Z}, \vb{\mathcal{B}})\mathbf{\Xi}}} ^2_{F} + \lambda\norm{\mathbf{\Xi}}_1,
     \label{eq: sparse_regression}
\end{equation}
where $\mathbf{Z} = \left[\,\vb{z}\left(t_1, \bm{\beta}_1\right) \,|\, \ldots \,|\, \vb{z}\left(t_{N_t}, \bm{\beta}_{N_\beta}\right)\right]^T\in\mathbb{R}^{N_t N_\beta \times n}$ is the matrix of low-dimensional encoded snapshots, $\dot{\mathbf{Z}}=\dot{\tilde{\mathbf{X}}}\nabla^T_{\tilde {\mathbf{x}}}\bm{\varphi}(\tilde{\mathbf{X}})$ contains their time derivatives, $\vb{\mathcal{B}}=\left[\,\bm{\beta}_1(t_1)\,|\, \ldots \,|\, \bm{\beta}_{N_\beta}(t_{N_t})\,\right]^T\in \mathbb{R}^{N_t N_\beta \times p}$ collects the parameter/forcing data, while $\mathbf{\Theta}$ is intended to be applied row-wise to the input matrices. Here, $\norm{\mathbf{\Xi}}_1$ is a regularization term chosen to promote sparsity in $\vb{\Xi}$ and $\lambda \in \mathbb{R}^+$ weights its contribution. \paolo{More sophisticated techniques, such as, e.g., sequentially thresholded least-square algorithm, might be used to promote sparsity on the entries of $\vb{\Xi}$ more efficiently \cite{champion2019data, bakarji2022discovering}}. Prior knowledge can be incorporated by imposing constraints on the entries of $\vb{\Xi}$, which reflect physical properties of the problem. \paolo{Specific entries can be masked and constrained to assume values of given coefficients. As a result, they remain constant over the training and only unconstrained entries of $\vb{\Xi}$ are updated at each iteration.} In the following, we discuss how to practically solve the sparse regression problem \eqref{eq: sparse_regression} together with \eqref{eq: autoencoder_loss}.




\subsection{Offline training}
\label{sect: offline_training}

\noindent In the previous sections we introduced the overall framework which consists of \textit{(i)} a dimensionality reduction and a change of coordinates from physical to intrinsic ones via POD and AE and \textit{(ii)} the identification of the latent dynamics by sparse regression. Tasks \textit{(i)} and \textit{(ii)} are addressed by solving minimization problems \eqref{eq: autoencoder_loss} and \eqref{eq: sparse_regression}, respectively. Although of different types, these tasks are intrinsically related \cite{champion2019data, bakarji2022discovering}. Indeed, the possibility of finding an accurate latent dynamical description with few meaningful and interpretable terms depends strongly on the choice of intrinsic coordinates, thus on the AE mapping.

\begin{figure}[t!]
    \centering
    \includegraphics[width=0.975\linewidth]{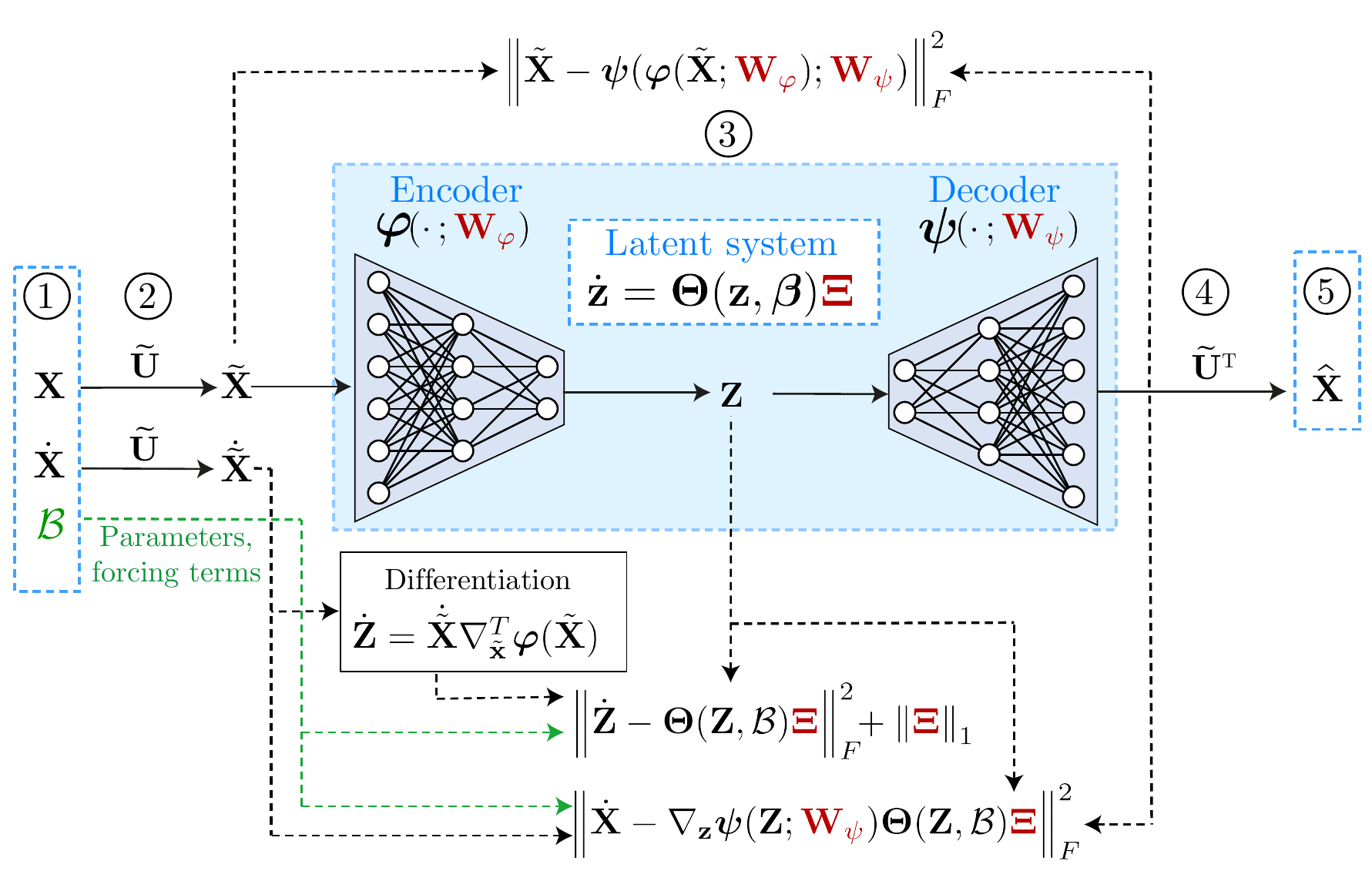}
    \caption{Neural network used during the training stage. The different stages of the training procedure are schematized as follows.  $\bfX$ and  $\dot{\bfX}$  are provided as input
    data (1) and  are fed to the POD linear reduction procedure (2). The POD coordinates $\tilde{\bfX}$ are processed by the AE+SINDy (3), where the AE extracts the latent coordinates $\mathbf{Z}$ and these, together with $\dot{\bf{Z}}$ and $\vb{\mathcal{B}}$, are used to compute the SINDy approximation of the latent dynamics. A linear POD reconstruction (4) provides the final output (5). The trainable parameters are highlighted in red.}
    \label{fig: training}
\end{figure}

Therefore, both tasks are performed by a single neural network architecture model denoted AE+SINDy, as illustrated in Fig.~\ref{fig: training}. This unified structure allows us to reformulate \eqref{eq: autoencoder_loss} and \eqref{eq: sparse_regression} as a single minimization problem by incorporating the latter in the former, namely the sparse regression terms are inserted as additional weighted components in the AE loss function. In this perspective, the unknown entries of $\mathbf{\Xi}$ are treated as network variables and they are estimated at the same time as AE parameters $\mathbf{W}_\varphi$ and $\mathbf{W}_\psi$ during the training of the neural network. Specifically, the optimization problem reads as
{\small{
\begin{equation}
     \begin{array}{ll}
     & \underbrace{\norm{\tilde{\mathbf{X}} - \bm{\psi}\left(\bm{\varphi}\left(\tilde{\mathbf{X}};\mathbf{W}_\varphi\right);\mathbf{W}_\psi\right)}^2_{F}}_\text{\small Autoencoder loss} + \underbrace{\lambda_1\norm{{\mathbf{\dot{Z}} - \mathbf{\Theta}(\mathbf{Z},\vb{\mathcal{B}})\mathbf{\Xi}}} ^2_{F}+ \lambda_2 \norm{\mathbf{\Xi}}_1}_\text{\small Sparse regression loss}  \\
     & \hspace{4cm} + \underbrace{\lambda_3 \norm{\dot{\mathbf{X}} - \nabla_\mathbf{z}\psi(\mathbf{Z};\mathbf{W}_\psi)\mathbf{\Theta}(\mathbf{Z},\vb{\mathcal{B}})\mathbf{\Xi}}_F^2}_\text{\small Consistency loss} \rightarrow \underset{{\mathbf{W}_\varphi, \mathbf{W}_\psi, \,\mathbf{\Xi}}}{\min},
     \end{array}
     \label{eq: loss}
\end{equation}
}}
\noindent where the latter term, denoted as \textit{consistency loss}, ensures that the time derivative of the network output matches the input time derivatives $\dot{\mathbf{X}}$. This term is proposed in \cite{champion2019data} and extended here to account for parametric/forcing dependencies. The coefficients $\lambda_1, \lambda_2,\lambda_3 \in \mathbb{R}^+$ are hyperparameters which weight the contribution of each term. \paolo{As a rule of thumb, the AE loss represents the leading term, so the coefficient $\lambda_1$, relative to the sparse regression term, should be smaller than unity and kept two orders of magnitude larger than 
$\lambda_2$ and $\lambda_3$, 
these latter weighting regularization and consistency terms, respectively. More precise values need to be adjusted to the specific dataset at-hand, being the choice of these coefficients problem-dependent.}

\paolo{A further hyperparameter to be tuned is the latent dimension $n$. This can be set equal to the dimension of the low-dimensional manifold, if this information is available. Otherwise, the behavior of the \textit{autoencoder loss} \eqref{eq: autoencoder_loss} as function of the number of latent variables may be used as an investigation tool to find the minimal latent dimension $n$. This value is the one for which further reducing the number of latent variables results in a drastic increase of the error. Although no examples are shown in the present work, AE+SINDy can be also employed by setting a number of variables larger than the dimension of the manifold. This may require placing more emphasis on \textit{sparse regression loss} in \eqref{eq: loss}, since the system identification task becomes more complex as both the number of latent equations and the number of features in the SINDy library increase.}

The \textit{offline training} procedure consists in the minimization of \eqref{eq: loss} through backpropagation with ADAM algorithm \cite{kingma2014adam}.
Once the joint training is finished, we freeze the AE weights $\mathbf{W}_\varphi, \mathbf{W}_\psi$ and we fine-tune just the SINDy coefficients $\vb{\Xi}$ for a better estimation of latent model coefficients \cite{bakarji2022discovering}.

\subsection{Online testing}
\label{sect: online_testing}
\begin{figure}[t!]
\centering
\includegraphics[width=0.99\linewidth]{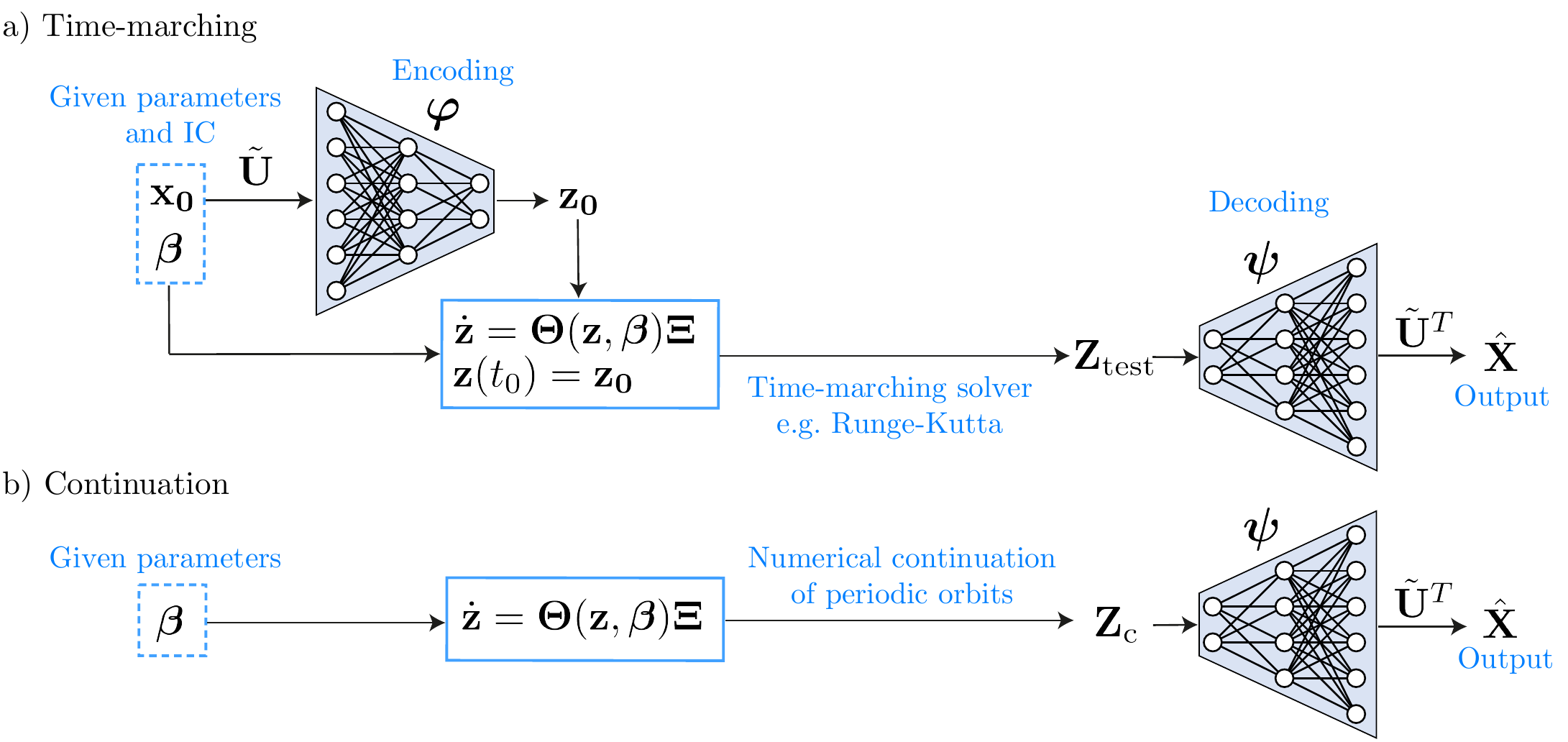}
\caption{Schematic representation of the setup used during the testing stage with the two approaches considered: time-marching solution and continuation of periodic orbits. 
In the former case, for a given value of the parameters $\vb{beta}$, the ROM solution is computed through a time marching scheme, like, e.g., Runge-Kutta methods, 
starting from a given initial condition. The computed ROM solution spanning the time range $t\in[t_0:t_\text{end}]$ is passed to the decoder.
In the latter approach, the numerical continuation computes the periodic orbits of the ROM within a set of parameters. Latent orbits are collected in $\mathbf{Z}_\text{C}$ and then decoded.
}
\label{fig: testing}
\end{figure}

\noindent Once the AE+SINDy model is trained \textit{offline}, we can query it \textit{online} to efficiently compute the entire time evolution of the full order system for different initial conditions and new parameter and forcing instances.
The procedure, illustrated in Fig.~\ref{fig: testing}a , consists of the following steps:
\begin{enumerate} 
    \item[1.] \textbf{encoding}. Once the encoder $\bm{\varphi}$ is trained, it is employed as mapping from the physical, full order space to the latent one. Hence, the physical initial condition $\vb{x_0}\in \mathbb{R}^{N}$ is encoded in order to obtain the latent counterpart $\vb{z_0} = \bm{\varphi}(\vb{x_0}\tilde{\vb{U}})\in \mathbb{R}^n$;
    
    \item[2.] \textbf{time-marching ROM solver}. SINDy feature library $\vb{\Theta}$ is populated with parameter and forcing terms $\bm{\beta}$, such that -- together with the feature coefficients $\vb{\Xi}$ identified during the training -- it defines the right-hand-side of the latent ODE dynamical system $\dot{\vb{z}} = \vb{\Theta}(\vb{z},\bm{\beta})\vb{\Xi}$. 
   This equations are numerically integrated, e.g. by Runge-Kutta time-marching scheme, starting from $\vb{z}(t_0)=\vb{z_0}$ up to an arbitrary time $t_\text{end}$ -- which may be much longer than the final time employed in the training phase.
    As result, we obtain the temporal evolution of the latent variables, collected in the matrix $\vb{Z_\text{test}}\in \mathbb{R}^{N_t^\text{test}\times n}$, where $N_t^\text{test}$ is the total number of testing time instants;
    
    \item[3.] \textbf{decoding}. The integrated latent variables are passed through the decoder and multiplied by the POD modes in order to reconstruct the entire time evolution of the original variables in the physical space, so $\hat{\vb{X}} = \bm{\psi}(\vb{Z}_\text{test})\tilde{\vb{U}}^T\in \mathbb{R}^{N_t^\text{test}\times N}$.
\end{enumerate}

These steps are extremely efficient from a computational standpoint: indeed, steps (1) and (3) only require the evaluation of pre-trained neural network functions (encoder and decoder, respectively); step (2) consists instead of the numerical integration of a low-dimensional system, a task which is much cheaper and faster than the resolution of high-dimensional numerical schemes for full order problems.

\subsection{Continuation of periodic orbits}
\label{sect: continuation}

\noindent As commented in Sect. \ref{sec:introduction}, one of the major benefits behind the identification of an explicit, low-dimensional, parameter-dependent latent dynamical system is the possibility to use continuation algorithms to track the evolution of periodic responses as a function of the system parameters and to perform 
a bifurcation analysis of the solutions  in order to have a complete portrait of the dynamics. 
As schematically represented in Fig.~\ref{fig: testing}b, once the continuation parameter is defined periodic orbits are computed in the latent space, collected in a  matrix $\mathbf{Z}_\text{C}$
and the corresponding full field solutions are eventually reconstructed with the decoder.

\begin{figure}[b!] 
\centering
\includegraphics[width=0.99\linewidth]{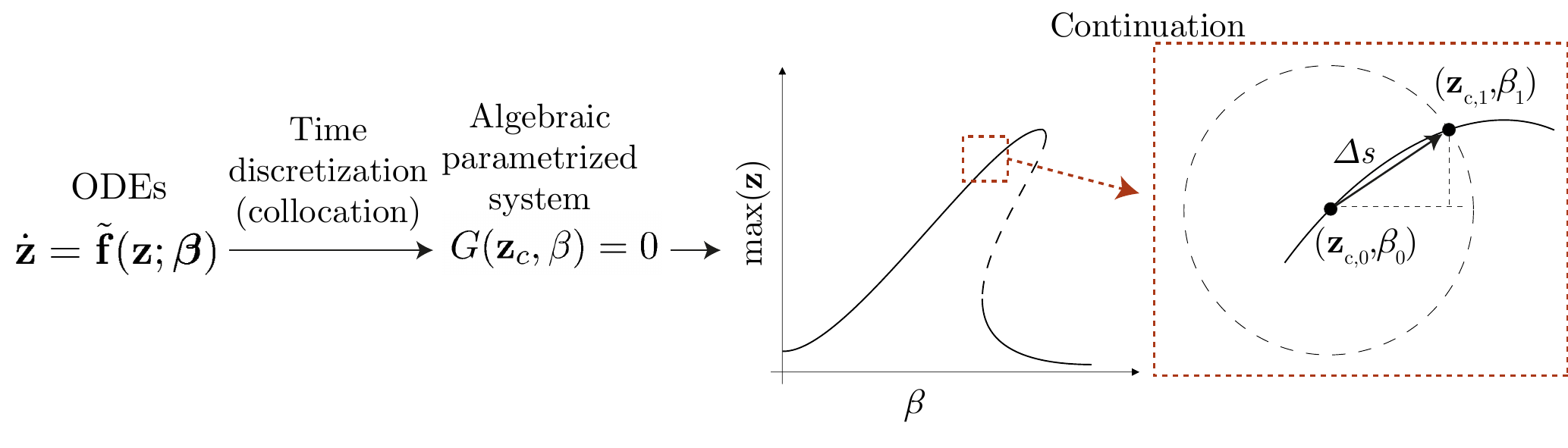}
\caption{
Sketch of the continuation of periodic orbits. 
Starting from a system of first order ODEs the system response is discretized in time using collocation methods. 
The continuation approach introduces a distance constraint between 
successive solutions corresponding to varying $\beta$. 
This allows to follow both stable as well as unstable branches.
}
\label{fig:cont}
\end{figure}


Periodic responses can be directly computed with several numerical strategies like harmonic balance \cite{krack2019harmonic}, collocation \cite{doedel2007auto}, and shooting methods \cite{osborne1969shooting}.
Focusing e.g.\ on the simplest version of collocation methods, the latent variables $\vb{z}$ 
are expressed over one period, of duration $\tau$, as the weighted sum of predefined time-shape functions
$N_i$ as
\begin{align}
\label{eq:continuation}
\vb{z}(t)=\paolo{\sum_{i=1}^{n_c}} N_i(t)\vb{z}_{c,i}  \quad t\in(0,\tau)
\end{align}
with the additional constraint $\vb{z}(0)=\vb{z}(\tau)$.  \paolo{$\vb{z}_{c,i}$ denotes the value of $\vb{z}$ at the $i-th$ collocation point and $n_c$ is the total number of collocation points.} The shape functions
might be defined globally over the period or locally, resorting to a partition
of the period into finite elements in time.
%
Inserting Eq.\eqref{eq:continuation} into Eq.\eqref{eq: dy_eq_reduced}, this latter transforms in a nonlinear system of algebraic equations of the form:
\begin{equation}
     G({\vb{z}_c},\beta)=0
     \label{eq:Gcont}
\end{equation}
where ${\vb{z}_c}$ represents the collection of the \paolo{$\{\vb{z}_{c,i}\}_{i=1}^{n_c}$ values } in Eq.\eqref{eq:continuation}
and $\beta$ is the selected continuation parameter, i.e.\ the parameter which is
varied in a predefined range over which the solutions for $\vb{z}_c$ are sought.
For instance, in the numerical applications discussed next,
the continuation parameters will be the actuation frequency for the clamped-clamped beam
and the Reynolds number for the simulation of the flow past a cylinder. 
This equation is complemented by suitable phase conditions, see e.g. \cite{krauskopf2007numerical}, to guarantee uniqueness.

Let us now suppose that $\vb{z}_c^{(k)}$ is a known solution of the system \paolo{in correspondence of the parameter $\beta^{(k)}$}. 
The simplest choice consists in taking $\beta$ as continuation parameter, fixing 
$\beta^{(k+1)}=\beta^{(k)}+\Delta\beta$ and solving \eqref{eq:Gcont}
for $\vb{z}_c^{(k+1)}$ through an iterative Newton–Raphson procedure. 
However, this procedure fails e.g.\ in the presence of an unstable branch as in Fig.~\ref{fig:cont}. 
Indeed, by imposing an increment $\Delta\beta > 0$ at the peak, 
the solution would jump to stable solutions on the lower branch, 
completely missing the unstable dashed branch. 
For this reason, it is customary to
introduce an arc-length control in which $\Delta\beta$ is part of the unknowns 
and the abscissa $s$ along the solution curve is taken as the continuation parameter. 
A new constraint is added, a typical choice being:
\[
F(\Delta\vb{z}_c, \Delta\beta) = (\Delta\vb{z}_c)^T\Delta\vb{z}_c + (\Delta\beta)^2 -(\Delta s)^2 = 0
\]
where $\Delta s$ is the user defined {\it distance} between successive solutions.
An alternative is the Keller’s pseudo arc-length method \cite{krauskopf2007numerical} in which the increment 
$(\Delta\vb{z}_c, \Delta\beta)$
such that its projection along a specific direction (typically the tangent to the 
$\vb{z}_c,\beta$ manifold) has length $\Delta s$.
%
These methods are implemented in many ready-to-use packages like \texttt{Auto07p} \cite{doedel2007auto}, that implements collocation methods in \texttt{FORTRAN} to perform numerical continuation and bifurcation analysis; \texttt{Manlab}, a \texttt{Matlab} tool that uses HB methods and Asymptotic Numerical Method \cite{guillot2019taylor,guillot2020purely}; \paolo{\texttt{PyCont}\cite{PyCont} package on \texttt{Python}; }
\texttt{Nlvib}, that also exploits HB methods \cite{krack2019harmonic}, and many others among which we mention 
\texttt{COCO} \cite{dankowicz2013recipes} and \texttt{BifurcationKit} \cite{veltz2020bifurcationkit}. 
The numerical examples addressed in this work
will be solved using the Matlab \texttt{MATCONT} package \cite{dhooge2006matcont}, that exploits collocation methods to perform the continuation of periodic orbits.

\section{Applications}
\label{sec:applications}
\subsection{Hardening behaviour of a clamped clamped beam}
\label{sec:beam}
\noindent To highlight the capabilities of the AE+SINDy approach, we first address a structural mechanics problem in which a straight beam MEMS resonator is excited at resonance. 
Despite its simplicity, such a problem has practical applications like in micro resonators \cite{zega2020numerical}, where geometric nonlinearities provide meaningful contributions to the system dynamic response.
\subsubsection{Problem description}
\begin{figure}[b!]
    \centering
    \includegraphics[width=0.8\linewidth]{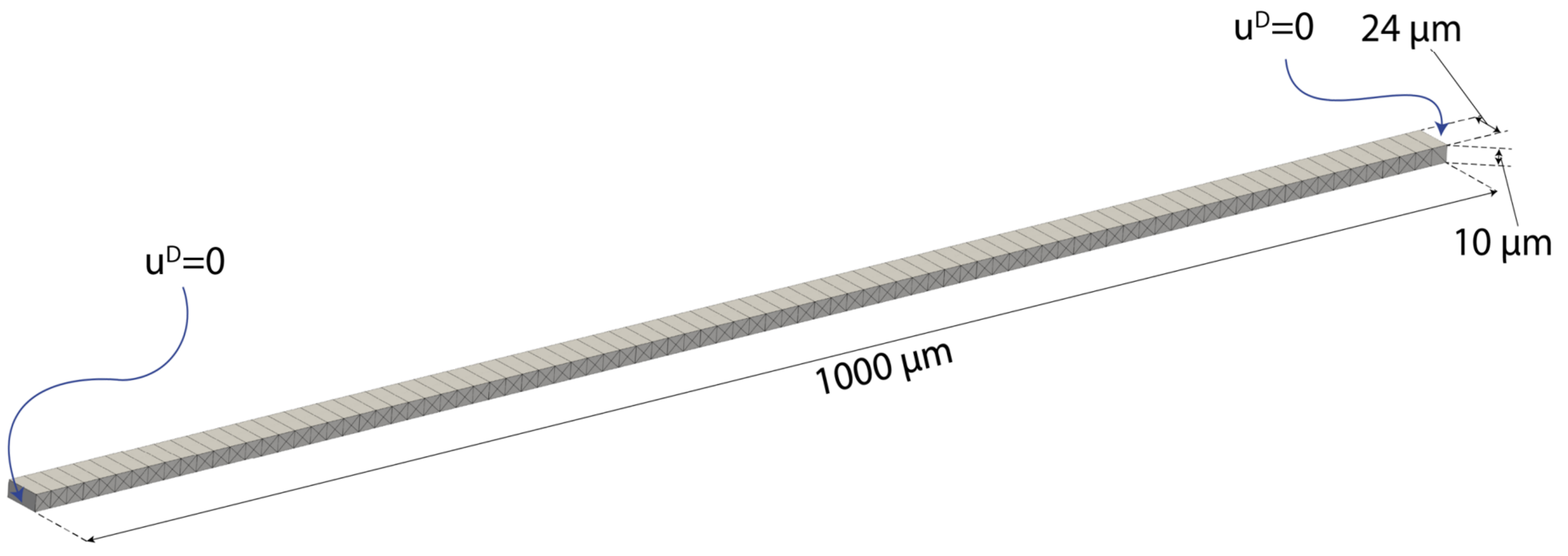}
    \caption{Schematic representation of the beam with the mesh used in the FOM simulations. Dirichlet boundaries are highlighted.}
    \label{fig: beam}
\end{figure}

\noindent We consider the doubly clamped beam depicted in Fig.~\ref{fig: beam} having length $L=1000$\micr with a rectangular cross-section of dimensions 10\micr$\times$24\micr, made of isotropic polysilicon \cite{corigliano2004mechanical}, with density $\rho=2330$\,Kg/m$^3$, Young modulus $E=167$\,GPa and  Poisson coefficient $\nu=0.22$. 
The first bending eigenfrequency is $\omega_0=0.5475\,\text{rad}/\mu\text{s}$

{In the undeformed configuration}, the device occupies the domain $\Omega_0$ described by material coordinates $\bm{x}$.
The boundary $\partial \Omega_0$ is partitioned in $\partial\Omega_D$ and  $\partial\Omega_N$ where homogeneous Dirichlet and Neumann boundary conditions are enforced, respectively. 
The external excitation is here provided by fictitious time-periodic body forces $\bfB(\bm{x},t)$.
The governing system of PDEs is classical in the context of solid mechanics problems in large transformations \cite{malvern1969introduction} and is formulated in terms of the displacement field $\bfu$ as follows:
\begin{subequations} 
	\label{eq:strongmech}
	\begin{align}
		\rho_0 \ddot{\bfu}(\bm{x},t) - \nabla\cdot\bfP(\bm{x},t)- 
		\rho_0 \bfB(\bm{x},t;\bfmu)=\bfzero, \; 
		&   \quad (\bm{x},t)\in \Omega_0 \times (0,T),
		\label{eq:strongmech_a}
		\\
		\bfP(\bm{x},t)\cdot\bfN(\bm{x})= \bfzero,\; 
		& \quad (\bm{x},t)\in \partial \Omega_N \times (0,T),
		\label{eq:strongmech_c}
		\\
		\bfu(\bm{x},t)=\bfzero,\; 
		& \quad (\bm{x},t)\in \partial \Omega_D \times (0,T). 
		\label{eq:strongmech_d}
	\end{align}
\end{subequations}

%

Eq.~\eqref{eq:strongmech_a} expresses the conservation of momentum
where $\rho_0$ is the initial density  and $\bfP$ is the first Piola-Kirchhoff stress.
The device is made of cubic single crystal silicon or polysilicon, thus admitting only small strains,  a condition which is well described by the Saint Venant-Kirchhoff constitutive model $\bfS(\bm{x},t)=\cA(\bm{x}):\bfE(\bm{x},t)$ where $\bfS$ is the second Piola-Kirchhoff stress tensor, $\cA$ is the fourth-order elasticity tensor and $\bfE(\bm{x},t)=\frac{1}{2}\left(\nabla\bfu(\bm{x},t)+\nabla^T\bfu(\bm{x},t)+\nabla^T\bfu(\bm{x},t) \cdot\nabla\bfu(\bm{x},t)\right)$ is the Green-Lagrange strain tensor.
Eq.~\eqref{eq:strongmech_c} and \eqref{eq:strongmech_d} define the 
Neumann and Dirichlet boundary conditions respectively. 
%
%
Within the context at hand, it is worthy to highlight that Eq.~\eqref{eq:strongmech} exactly accounts for geometric (elastic and inertia) nonlinearities, e.g., large rotations or nonlinear mode coupling. The spatial discretization of Eq.~\eqref{eq:strongmech}, e.g., by means of the finite element method, with the additional inclusion of a Rayleigh model damping term, yields a FOM under the form of a system of coupled first-order nonlinear differential equations that reads as:
\begin{subequations} 
	\label{eq:PPV_d2}
	\begin{align}
		&\bfM \dot{\bfv}_h(t) + \bfC\bfv_h(t)  + \bfK\bfu_h(t) + \bfG(\bfu_h,\bfu_h) + \bfH(\bfu_h,\bfu_h,\bfu_h) - 
		\bfF(t;\bfbeta)={\bf 0}, & \quad t \in (0,T) \\
		&\dot{\bfu}_h(t) - \bfv_h(t)  = {\bf 0}, & \quad t \in (0,T)   \\
		& {\bfu}_h(0) = {\bf 0}, \ \ \bfv_h(0)  =  {\bf 0},
	\end{align}
\end{subequations}
%
where the vectors $\bfu_h(t), \bfv_h(t) \in \mathbb{R}^{N_h}$ collect the $N_h$ unknown displacements and velocity nodal values respectively, $\bfM \in \mathbb{R}^{N_h \times N_h}$ is the mass matrix, $\bfC=(\omega_0/Q)\bfM$ is the Rayleigh mass-proportional damping matrix with quality factor $Q=50$. The internal force vector has been exactly decomposed in linear, quadratic, and cubic power terms of the displacement: $\bfK \in \mathbb{R}^{N_h\times N_h} $ is the stiffness  matrix related to the linearized system, while $\bfG \in \mathbb{R}^{N_h}$ and $\bfH \in \mathbb{R}^{N_h}$ are  vectors related to second- and third-order terms, respectively, \paolo{and indicated with the notation $ \bfG(\bfu_h,\bfu_h)=\sum_{k,l}\bfG_{kl}{u_h}_k{u_h}_l$ and $ \bfH(\bfu_h,\bfu_h,\bfu_h)=\sum_{k,l,m}\bfH_{klm}{u_h}_k {u_h}_l {u_h}_m$, where $\bfG_{kl}$ stands for the vector of coefficients $G^i_{kl}$, for $i=1,\ldots,N_h$, and analogously $\bfH_{klm}$ is a vector of coefficients   $H^i_{klm}$. We refer to \cite{opreni2022high} for a detailed description of these terms.}
$\bfF(t;\bfbeta) \in \mathbb{R}^{N_h}$ is the nodal force vector which depends on the vector of parameters $\bfmu$. 
In this application the device is excited, for simplicity, by a body load such that 
$\bfF=\bfM \bfphi_1 F \cos(\omega t)$ is
proportional to the first eigenmode $\bfphi_1$,  with $\bfM$ mass matrix and $F$ load multiplier.
In the application at hand the input parameter vector hence becomes $\bfbeta=[\omega,F] \in\mathcal{P}$, with $\mathcal{P}$ a closed and bounded set of dimension $2$.  

We point out that, independently of the mesh size, 
the solution of the FOM \eqref{eq:PPV_d2}  is usually a challenging task since MEMS devices feature high quality factors while only the steady-state response is of practical interest \cite{opreni2021analysis}. 
Indeed, the performance of such devices is typically characterized by means of the so called Frequency Response Functions (FRFs)
(see e.g. Figure~\ref{fig: beam_rec}) which, 
for different values of the load multipliers, 
express a steady-state output quantity of interest, like the maximum deflection, 
for all the values of angular frequency $\omega$ of the forcing within a prescribed range.
As a consequence, the generation of reliable and efficient ROMs for this type of applications has recently stimulated intensive research
with both standard and deep learning approaches; see, e.g., \cite{opreni2021model,gobat2022virtual,fresca2022deep,fresca2022pod}.


\subsubsection{Dataset}
\noindent We consider 2 values for the amplitude, $F\in\{0.125,0.250\}\,\mu\text{N}$, 
and 28 values for the frequency $\omega$ in the range $[0.526, 0.564]\,\text{rad}/\mu\text{s}$ with a finer sampling around the natural frequency $\omega_0$. Hence 
the total number of instances is $N_\beta = |\mathcal{P}| = 56$. The parameter set $\mathcal{P}$ is randomly partitioned in training and testing subsets, $\mathcal{P}_\text{train}$ and $\mathcal{P}_\text{test}$, such that $N_\beta^\text{train} = |\mathcal{P}_\text{train}| = 54$ and $N_\beta^\text{test} = |\mathcal{P}_\text{test}| = 2$.\\
For each choice of training parameters, $(\omega, F)\in \mathcal{P}_\text{train}$, 
we numerically approximate the FOM solutions for displacement and velocity up to time $T = 390\,{\mu}s$ 
and we collect them in the matrices $\mathbf{X}_{\bfbeta}, \dot{\mathbf{X}}_{\bfbeta} \in \mathbb{R}^{N_t \times N}$, where $N_t = 5000$ and $N = 7821$ are the number of time-steps and spatial degrees of freedom of the FOM, respectively, obtained by considering a mesh with 2607 nodes and linear finite elements. The matrices $\{\mathbf{X}_{\bfbeta}\}_{\bfbeta \in \mathcal{P}_\text{train}}$ are then stacked together, obtaining a single matrix $\mathbf{X}_\text{train}\in\mathbb{R}^{N_t N_\beta^\text{train}\times N}$, which contains the displacement snapshots for all the training parameters.
Analogously to $\mathbf{X}_\text{train}$, we fill a matrix $\dot{\mathbf{X}}_\text{train}$ with the corresponding velocity snapshots.

\paragraph*{Preprocessing} We preliminary reduce the dimensionality of the system by POD, moving from a dimension of $N = 7821$ to $N_\text{POD}=64$. In order to do so, first a reduced basis is constructed by applying POD on the training displacement snapshot matrix $\mathbf{X}_\text{train}$. Next, all the FOM data are projected onto the first $N_\text{POD} = 64$ bases, yielding POD coordinates denoted as $\tilde{\mathbf{X}}_\text{train},\dot{\tilde{\mathbf{X}}}_\text{train} \in \mathbb{R}^{N_t N_\beta^\text{train}\times N_\text{POD}}$. 

Moreover, the neural network optimization algorithm is more efficient if the training data is in a limited magnitude range, typically in the order of unity, as the standard nonlinear activation functions perform at best in this range. Hence, 
we chose to rescale each of the 64 input features with respect to its maximum in absolute value, such that they all fall in the range $[-1,1]$.

\subsubsection{AE+SINDy architecture}
\noindent The encoder block of the AE is composed of 3 hidden dense layers consisting of 64, 32 and 16 neural units. The decoder has a symmetrical structure.  The dimensionality of the latent space corresponds to the size of the encoder output (as well as that of decoder input) and it is set to 1. 
Indeed, as discussed in \cite{fresca2022deep,fresca2022pod,gobat2022virtual},
the dynamics underlying the system lies on a two dimensional manifold in the phase space \cite{opreni2021model}, but the dependence on the velocity is minimal for the master bending mode that we inspect herein.
Therefore, we aim to identify a latent dynamical system of the form
\begin{equation}
\ddot{z} = f(z, \dot{z}, \bm{\beta}),
\label{eq: second_order}
\end{equation}
where $z$ is the latent variable, $\dot{z}$ and $\ddot{z}$ are its first- and second-order time derivatives, respectively, and $\bm{\beta}$ accounts for the parameter and forcing terms. All variables are time-dependent, although the dependency is not explicit for conciseness.

In order to estimate $f$ by SINDy, we rewrite the second-order ODE \eqref{eq: second_order} as a system of two first-order ODEs, 
\begin{equation}
    \begin{aligned}
        \dot{z} &= y,\\
        \dot{y} &= f(z,y,\bfbeta),
    \end{aligned}
    \label{eq: beam_second_order}
\end{equation}
which can be easily expressed in a suitable form for the SINDy framework, namely
\begin{equation}
    \dot{\mathbf{z}} = \mathbf{f}\left(\mathbf{z}; \bm{\beta}\right)\approx \mathbf{\Theta}\left(\mathbf{z};\bm{\beta}\right)\mathbf{\Xi}.
    \label{eq: sindy_beam}
\end{equation}
Here, $\mathbf{\Theta}$ represents a library of polynomials up to the third degree with respect to $\mathbf{z} = \left[z, y\right]^T$ and of the first degree with respect to $\bm{\beta}$, while $\mathbf{\Xi}$ is the sparse unknown matrix of the multiplicative coefficients of the candidate features in $\mathbf{\Theta}$. 

Regarding the choice of $\bm{\beta}$, since the beam structure is forced harmonically with amplitude and frequency parameterized by $F$ and $\omega$, these are incorporated in the SINDy model through the library terms 
$F \cos(\omega t), F \sin(\omega t)$. 
Therefore, $\mathbf{\Theta}$ has the following form
\begin{equation*}
    \mathbf{\Theta}(\mathbf{z};\bm{\beta}) = \left[\, z\left(t\right) \,,\, y\left(t\right) \,,\, z^2\left(t\right) \,,\, z\left(t\right)y\left(t\right) \,,\,\ldots \,,\, y^3\left(t\right) \,,\, F \cos(\omega t) \,,\, F \sin(\omega t) \,\right] 
\end{equation*}

\noindent and contains $r=11$ candidate features. Furthermore, we remark that not all the entries of $\mathbf{\Xi} \in \mathbb{R}^{r \times n}$, with $n=2$,  need to be estimated. In fact, by construction, the first equation in \eqref{eq: beam_second_order} is known, thus the first column of $\mathbf{\Xi}$ -- which indeed represents the coefficients of this equation -- is all set to 0 except for the second entry, relative to $y$, which is set to 1. Moreover, we can incorporate prior-knowledge about the dynamics by setting the values of (or imposing constraints on) the coefficients of $\mathbf{\Xi}$. 
In structural problems governed by \eqref{eq: beam_second_order},  linear terms reduce to
$-\omega_0^2 z$ that just depends on the natural frequency $\omega_0$, which is known as the spectral properties of the linearized
system are the classical primal output of the FOM.
Thus, $\mathbf{\Xi}$ becomes
\[
\mathbf{\Xi} = \left[\begin{array}{cccccc}
     0 & 1 & 0 & \ldots & 0 & 0  \\
     -\omega_0^2 & \Xi_{2,2} & \Xi_{3,2} & \ldots & \Xi_{r-1,2} & \Xi_{r,2}
\end{array}\right]^T \in \mathbb{R}^{r \times n}.
\]
Following the procedure of Sect.~\ref{sect: offline_training}, the overall architecture (see Fig.~\ref{fig: training}) is trained on $\tilde{\mathbf{X}}_\text{train}$ and $ \dot{\tilde{\mathbf{X}}}_\text{train}$ to simultaneously reconstruct the POD coordinates with the AE as well as to learn the dynamics of the latent system by SINDy. 
The identified dynamical latent system finally reads: 
\begin{align}
\label{eq: pred_model}
\dot{z} =  &\, y,\nonumber 
\\
\dot{y} = & - 0.3 \,z - 0.011\, y + 0.003\, z^2 - 0.012\, y^2 - 0.113\, z^3 + 0.036\, z^2 y  \nonumber 
\\ 
& + 0.719\,zy^2 - 0.051 \,y^3 - 0.009 \,F \cos(\omega t).
\end{align}
%

\subsubsection{Results}
\noindent Once the offline training is concluded and the AE parameters $\mathbf{W}_\varphi$ and $\mathbf{W}_\psi$, as well as the SINDy coefficients $\mathbf{\Xi}$, are learned, we can employ the model for the online fast evaluation of new solutions or get insight into the dynamics by the continuation algorithms.

\begin{figure}[htp]
    \centering
\includegraphics[width=0.785\textwidth]{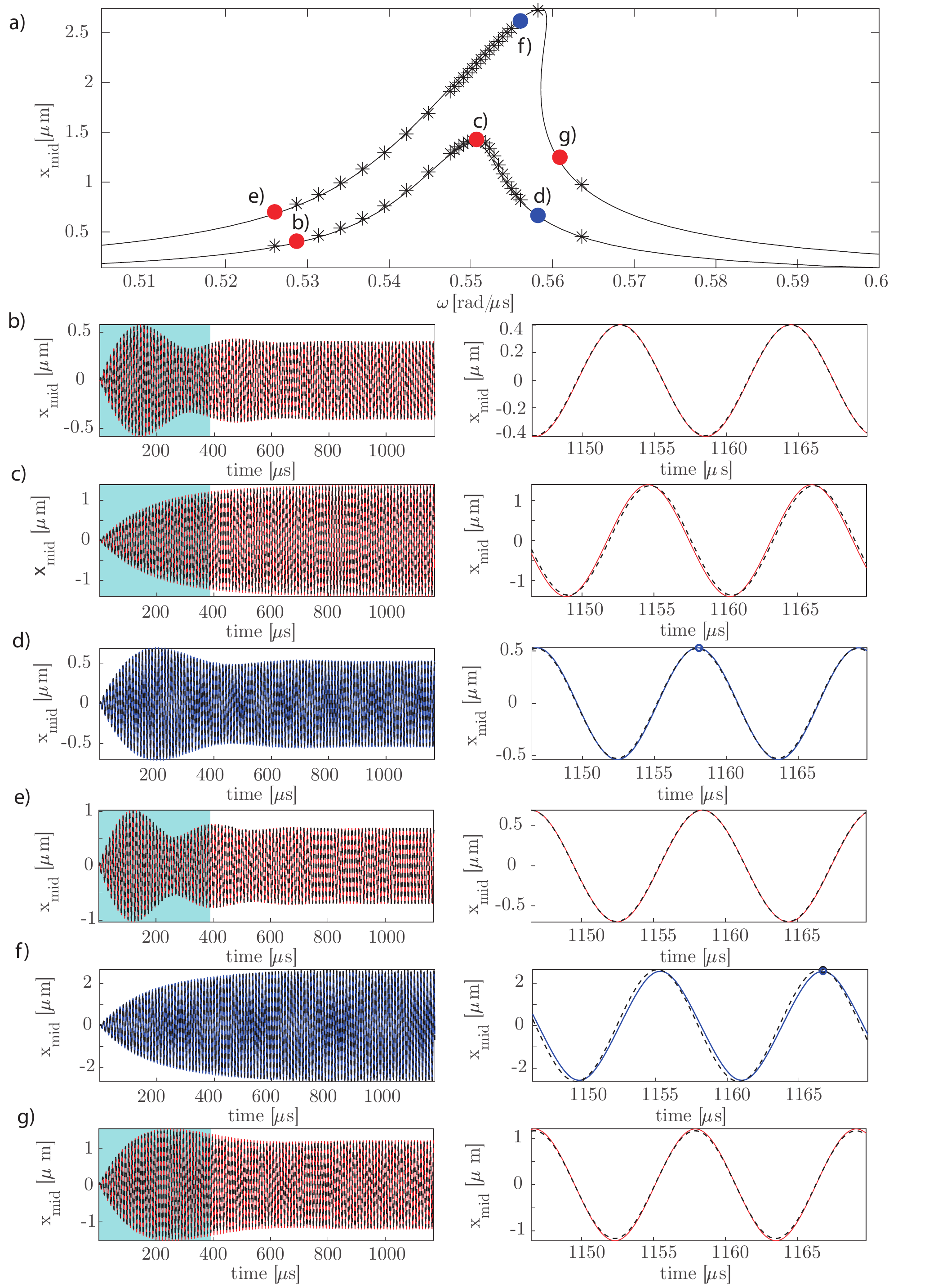}
    \caption{Comparison between the time evolution of the FOM numerical solution $\mathbf{X}$ and the predicted displacement $\hat{\mathbf{X}}$ on a specific mesh node positioned at mid-span of the beam, for different frequency-loading pairs.
    Fig.~a) FRF of the system. The continuous line denotes the FOM solution, while the markers represent the solutions obtained with the AE+SINDy after a long time integration of the ROM. The coloured markers correspond to solutions reported in subfigures b)-g). In particular, for each marker  the whole time history is plotted in the first column and an enlarged view of the last few periods is presented in the second column. The black dashed lines refer to the FOM solution, while the red and blue ones to the AE+SINDy solution on training and testing load levels, respectively. 
    The light blue shaded regions define the training time interval.
    The markers on Figs. d) and f) highlight the instants spatially reconstructed in Fig.\ref{fig:space_beam}.}
    \label{fig: beam_rec}
\end{figure}

\paragraph*{Time-marching solutions}

The trained model is used to efficiently compute solutions for new 
instances of the parameters, i.e.\ for the new forcing frequencies
$\omega$  and amplitude values $F$. In particular, we assess the performance on the test set $\mathcal{P}_\text{test}$. Following the procedure presented in Sect.~\ref{sect: online_testing}, initial conditions and forcing parameters are passed to the encoder and to the SINDy model respectively; then the identified latent system is integrated up to  $t_\text{end} = 1170\,{\mu}s $, i.e. three times longer than the training final time ($T = 390\,{\mu}s$).  In particular, at $t_\text{end}$ the system has reached the steady-state, while, during the training, only data covering a transient part of the dynamics have been fed to the network. Finally the physical solution is recovered via the decoder. In Fig.~\ref{fig: beam_rec} we compare the AE+SINDy predictions with the displacement approximated with the FOM. The model manages to approximate the FOM solution with great accuracy in the whole time window. In addition, long-time predictions for few training parameters are depicted in Fig.~\ref{fig: beam_rec} for a more comprehensive representation of the capacity of the method to extrapolate in time and to estimate the steady-state behavior in different regions of the parameter space.

These results highlight that \textit{(i)} the identified model is able to explain dynamical phenomena, such as the steady-state behavior,  while it has been trained only with snapshots covering a limited portion of the transient; \textit{(ii)} the AE+SINDy framework not only manages to forecast in long-time horizons but it \textit{simultaneously} generalizes with respect to new parameters; \textit{(iii)} the identified latent one-dimensional dynamics physically represent and accurately approximate the whole high-dimensional system.

\begin{figure}[t!]
    \centering
    \includegraphics[width=0.95\textwidth]{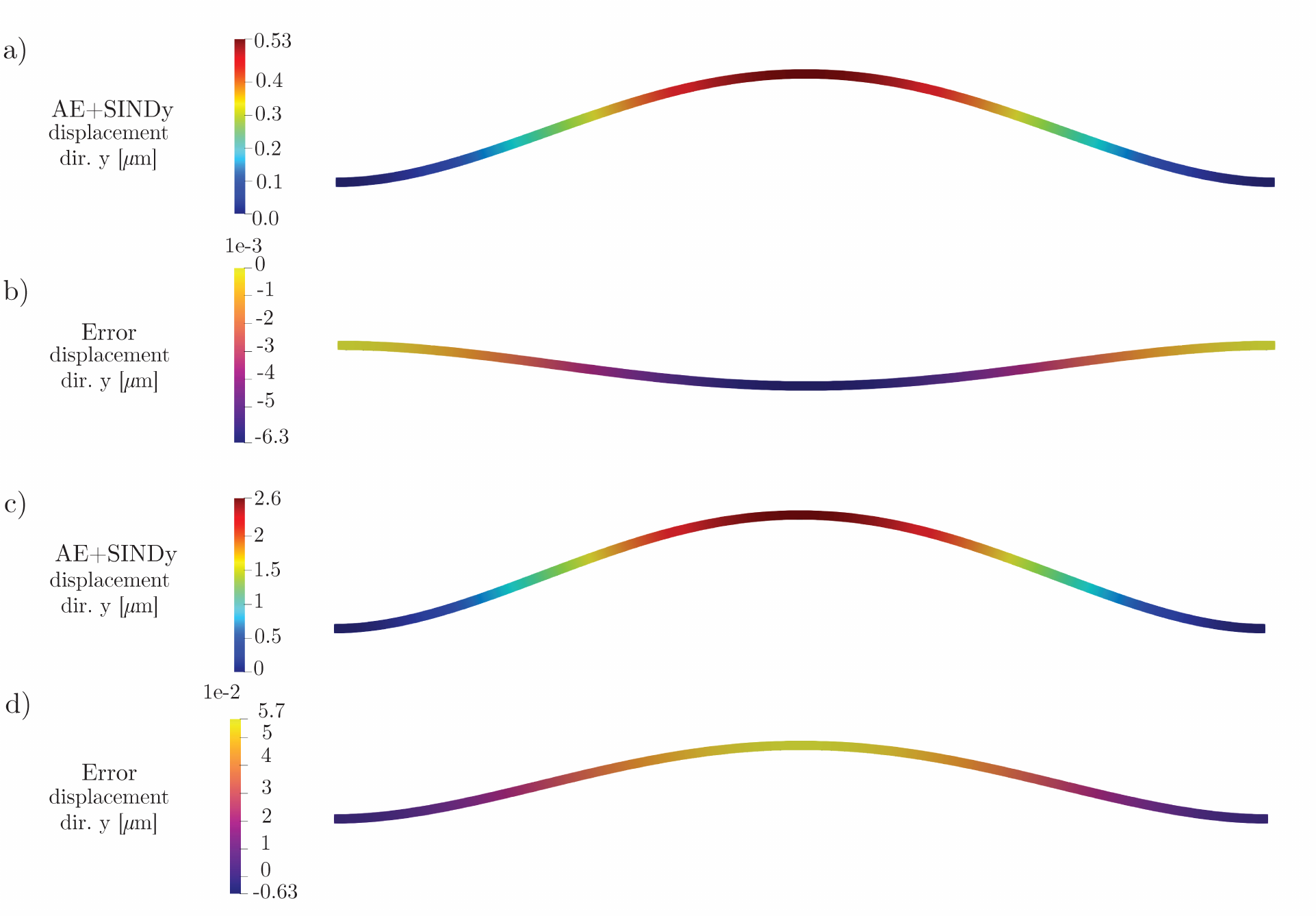}
    \caption{AE+SINDy spatial reconstruction of the displacement field on the y direction and the error with respect to the FOM. Fig. a-b) refer to the testing instance and time instant highlighted in Fig.~\ref{fig: beam_rec} d), while Fig. c-d) refer to the one in Fig.~\ref{fig: beam_rec} f).}
    \label{fig:space_beam}

\end{figure}


\paragraph*{Continuation of periodic orbits}

The low-dimensional latent dynamical system can be analysed 
with several different approaches and in particular with continuation algorithms that allow tracking directly the steady-state value of a quantity of interest 
as the parameters of the model are varied. 
In the problem at hand we are interested in plotting  FRFs, 
that are particularly meaningful because they summarize most of the dynamical information of the mechanical system in a compact form.

\begin{figure}[h!]
\centering
\includegraphics[width=0.99\textwidth]{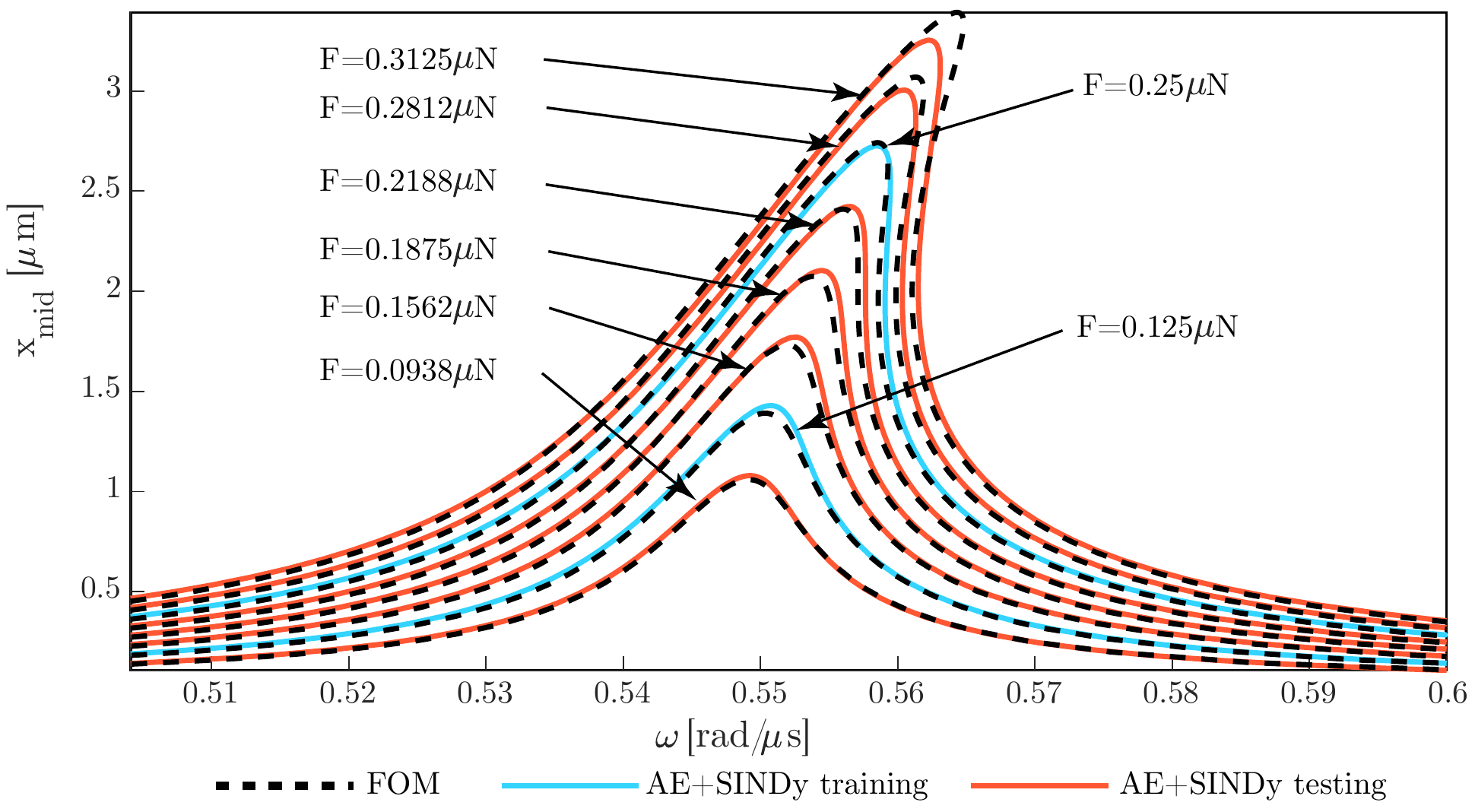} 
\caption{Comparison between FRFs obtained with the FOM \cite{opreni2021analysis} and the AE+SINDy. 
FOM solutions are plotted with dashed lines while AE+SINDy solutions with 
continuous ones. The red curves refers to testing load levels while the light blue one to training load levels. \paolo{FRFs are represented as continuous lines for illustrative clarity, since continuation algorithm provides data-points which are very close to each other.}}
\label{fig: FRF_beam}
\end{figure}

Following the continuation strategy presented in Sect.~\ref{sect: continuation}, steady-state solutions are computed in the latent space, then mapped, by passing them to the decoder, into the physical space, where finally FRFs are drawn. In Fig.~\ref{fig: FRF_beam} we illustrate the comparison between the predicted curves with the exact curves obtained from the FOM \cite{opreni2021analysis}.
Therefore, even though no steady-state information and unstable solutions have been provided during the training of the model, we still manage to track the evolution of the steady-state response changes with respect to the parameters, even in the unstable regions. Indeed, since the system is nonlinear we encounter bifurcations that affect the number of the solutions and their stability.

In Fig.~\ref{fig: FRF_beam}, we plot (in light blue) the FRFs corresponding to the training values  of 
$F$, i.e.\ $F\in\{0.125,0.250\}\,\mu\text{N}$. 
We note that the predicted curves follow with great accuracy the exact ones and, in particular, there is an excellent match even outside the span of training frequencies, that is  $[0.526, 0.564]\,\text{rad}/\mu\text{s}$, which denotes a remarkable capability of the AE+SINDy to extrapolate with respect to parameter $\omega$. 

Moreover, we plot (in red) FRFs for forcing amplitudes 
$F$ that have not been seen during the training phase. 
Also in this case, in general, there is a very good match with the exact curves, suggesting an excellent ability to generalize with respect to both the forcing amplitude $F$ and frequency $\omega$. 
A progressive degradation of the accuracy can be noted at the peak of FRFs for large values of forcing amplitudes outside the training range, e.g., for the curve corresponding to $F=0.3125\,\mu\text{N}$. This limitation in extrapolation is mainly due to low data sampling for the frequencies corresponding to peaks and to the fact that the AE performs a data-driven nonlinear coordinate transformation which may be less accurate when the nonlinearity content of the response increases well beyond what experienced during training. 
It is however worth stressing that far from the peaks the prediction is very accurate, even in the unstable branch of the FRFs.
In support of these remarks we further notice that, on the contrary, 
for the curve at $F = 0.0938\,\mu\text{N}$, i.e., below training range, we have a very accurate approximation.

\subsection{Fluid flow around a cylinder}
\label{sec:reynolds}
\noindent For this second application, our goal is to efficiently approximate the fluid velocity and pressure in the case of a fluid flow problem, at different regimes as the Reynolds number varies, as well as to empirically identify, through continuation algorithms, the bifurcation value of the transition from laminar to unsteady behaviour.


\subsubsection{Problem description}
\begin{figure}[b!]
    \centering
    \vspace{-0.2cm}
    \includegraphics[width = 0.9\textwidth]{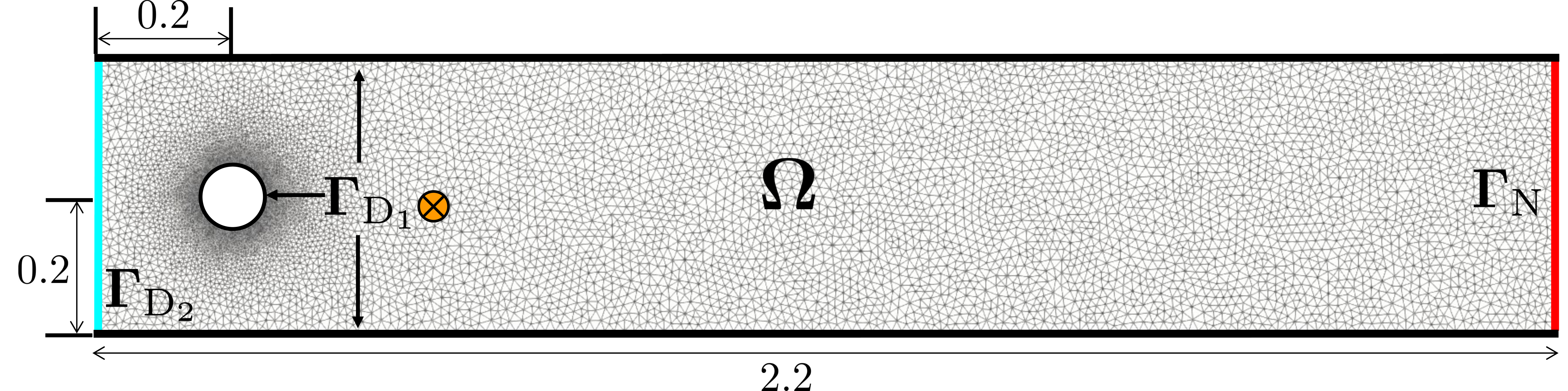}
    \caption{Geometry for the 2-D channel flow around a cylinder. All lengths are measured in meters. The node highlighted in orange is used as reference node for visualizations of results in Figs.~\ref{fig: tmp2},\ref{fig: continuation_rec}.}
    \label{fig:NS_geometry}
\end{figure}

\noindent We consider the well-known benchmark dealing with a two-dimensional fluid flow around a cylinder. The problem is described by the following unsteady Navier-Stokes equations for a viscous, incompressible Newtonian flow
\begin{equation}
    \begin{aligned}
           \rho \frac{\partial \textbf{v}}{\partial t} - \rho \textbf{v}\cdot \nabla\textbf{v} - \nabla \cdot \bm{\sigma}(\textbf{v},p) &=  \textbf{0}, \qquad &&(\bm{x},t)\in \Omega \times (0,T) \, ,  \\
    \nabla \cdot \textbf{v} &= 0, \qquad &&(\bm{x},t)\in \Omega \times (0,T) \, .
    \end{aligned}
    \label{eq: NS}
\end{equation}
Here, $\textbf{v}(\bm{x},t)$ and $p(\bm{x},t)$ represents respectively the velocity and pressure fields of the flow, $\rho = 1.0 \text{ kg/m}^3$ is the fluid density, $\bm{\sigma}(\vb{v},p)=-p\vb{I}+2\nu\bm{\epsilon}(\vb{v})$ is the stress tensor, $ \bm{\epsilon}(\vb{v})$ is the strain tensor, while $\nu$ is the kinematic viscosity. The domain $\Omega = (0, 2.2) \times (0, 0.41) \text{\textbackslash} B_r(0.2,0.2)$, with $r = 0.05$, consists in a bidimensional channel, where $B_r$ represents a cylindrical obstacle (see Fig.~\ref{fig:NS_geometry}). We prescribe the following boundary and initial conditions:
\begin{equation}
  \begin{aligned}
     \textbf{v} &= \textbf{0}, \qquad &&(\bm{x},t)\in \Gamma_{\text{D}_1} \times (0,T) \, , \\
    \textbf{v} &= \textbf{h}, \qquad &&(\bm{x},t)\in \Gamma_{\text{D}_2} \times (0,T)  \, , \\ 
    \bm{\sigma}(\textbf{v},p)\textbf{n} &= \textbf{0}, \qquad &&(\bm{x},t)\;\in \Gamma_\text{N} \times (0,T)  \, , \\ 
    \textbf{v}(\bm{x}, 0) &= \textbf{0}, \qquad &&\bm{x} \in \Omega \, ,
  \end{aligned}
\end{equation}
which represent a no-slip condition on $\Gamma_{\text{D}_1}$, a parabolic inflow of the form 
\[ \textbf{h}(\bm{x},t) = \left(\frac{4U(t)x_2(0.41-x_2)}{0.41^2},0\right)\,,  \qquad \text{where }
U(t)=\left\{
  \begin{array}{@{}ll@{}}
    0.75(1-\cos{(\pi t)}), & t <1 \\
    1.5, & t \geq 1
  \end{array}\right.
  \]
on the inlet $\Gamma_{\text{D}_2}$, open boundary conditions on the outlet $\Gamma_\text{N}$ and homogeneous initial conditions.



In this framework, $\nu = 1 / Re$ \cite{fresca2021real,conti2022multi}, where $Re$ is the Reynolds number, a quantity which allows to compare different flow regimes and characteristics. In the present study, we focus on the region $40 \leq Re \leq 60$. For the case at hand, when $Re < 49$ the flow is entirely laminar; instead, for larger Reynolds numbers, a pair of vortices are generated in the wake of the cylinder, periodically alternating between the top and bottom side, and the flow becomes unsteady \cite{zdravkovich1997flow, rajani2009numerical}.

For a set of values of the number of Reynolds in the range $\mathcal{P}=[40,60]$, we approximate the velocity components and pressure solutions up to $T=30 s$, for which the flow becomes fully developed. For the spatial and time discretization we consider finite elements  and a semi-implicit backward differentiation formulas (BDF) respectively, employing \textsc{Matlab} redbKIT library \cite{negri2016redbkit}.


The geometric mesh employed is shown in Fig.~\ref{fig:NS_geometry}, while the temporal discretization step is $\Delta t = 5 ms$. The total number of degrees of freedom is $N=73131$, obtained by considering quadratic finite elements for the velocity field and linear finite element for the pressure field, on a mesh made by 16478 triangular elements and 8239 vertices. Thus, numerical solutions of this type -- although extremely accurate -- are characterized by a very high dimensionality and they are computationally expensive and time consuming.  

Our goal is to exploit the AE+SINDy structure to compute the entire space-time solution for a new instance of the Reynolds number, by exploiting a ROM consisting of a low-dimensional dynamical system. In addition, we are interested in understanding whether the identified system can actually represent the change of fluid behaviour as the Reynolds number varies, in particular, if it manages to detect the bifurcation at the transition from laminar to unsteady flow (between $Re = 49$ and $50$, for this application \cite{zdravkovich1997flow, rajani2009numerical}).

\subsubsection{Dataset}
\noindent A uniform grid of $N_\beta = 21$ Reynolds number values over the parameter interval $\mathcal{P} = [40,60]$ is considered. For each parameter value, we compute numerical solutions for the pressure and the two velocity components of the fluid. For the sake of computational efficiency, we restrict our analysis to the time window $[t_0, T] = [15 s,30 s]$, by trimming off the first portion of the transient. $N_\beta^\text{train} = 19$ snapshots will be employed as training set and the remaining $N_\beta^\text{test} = 2$  for testing. In order to verify the accuracy of the proposed method in reconstructing different fluid behaviors, we consider as test instances the full order snapshots for $Re = \{43, 55\}$, which correspond to solutions in either the laminar or the unsteady regime, respectively. 

Training and testing snapshots are stacked respectively in matrices $\vb{X}_\text{train}\in\mathbb{R}^{N_t N_\beta^\text{train} \times N}$ and $\vb{X}_\text{test}\in\mathbb{R}^{N_t N_\beta^\text{test}\times N}$, where $N_t = 3000$ is the number of time-steps.

As in the previous example, a preliminary dimensionality reduction is performed by POD. In particular, POD is applied separately on the training snapshots of both pressure and velocities, {once they have been properly shifted to obtain zero mean snapshots}, since they feature different units of measurement. Then, for both velocity and pressure, we retain the projections of the snapshots onto the first 32 bases (for a final dimension of $N_\text{POD} = 64$). These POD coordinates are scaled by the square root of the corresponding singular value (to reemphasize low-energy POD modes) and divided by the overall maximum in absolute value. In this way, POD coordinates of velocity and pressure have the same order of magnitude (\i.e. unitary), which is also ideal for the neural network training. Finally, the two sets of POD coordinates are collected into a single matrix $\tilde{\vb{X}}_\text{train} \in \mathbb{R}^{N_t N_\beta^\text{train}\times N_\text{POD}}$. Here, time derivatives $\dot{\tilde{\vb{X}}}_\text{train}$ are computed numerically. Analogously, testing snapshots are projected onto the POD training modes, rescaled in the same way and collected in the matrix ${\tilde{\vb{X}}}_\text{test} \in \mathbb{R}^{N_t N_\beta^\text{test}\times N_\text{POD}}$.

\begin{figure}[p]
    \centering
    \includegraphics[width=1.\linewidth]{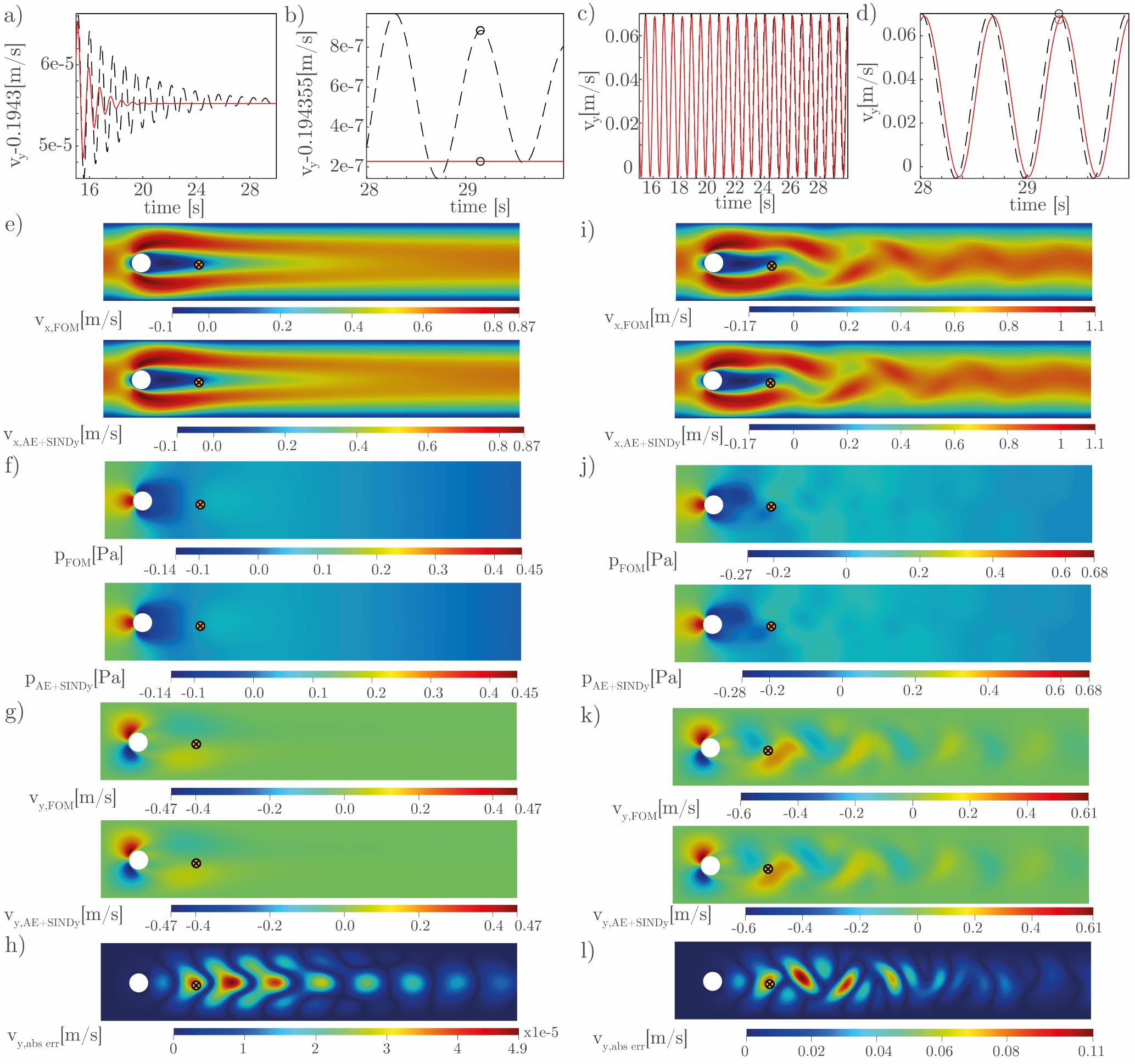}
    \caption{Time marching solutions: comparison between FOM and AE+SINDy. Fig.~a) time histories related to $Re=43$ \paolo{(AE+SINDy in red, FOM in black)}, enlarged view proposed in Fig.~b);
    Fig.~c) time histories related to $Re=55$, enlarged view proposed in Fig.~d);
    Fig.~e) FOM and AE+SINDy velocity field along the x direction with $Re=43$;
    Fig.~f) FOM and AE+SINDY pressure field with $Re=43$;
    Fig.~g) FOM and AE+SINDY velocity field along the y direction with $Re=43$;
    Fig.~h) absolute difference velocity field along the y between the FOM and AE+SINDy with $Re=43$; 
    Fig.~i)  FOM and AE+SINDY velocity field along the x direction with $Re=55$;
    Fig.~j) FOM and AE+SINDY pressure field with $Re=55$;
    Fig.~k) FOM and AE+SINDY velocity field along the y direction with $Re=55$;
    Fig.~l) absolute difference between the FOM and AE+SINDy with $Re=55$ on the y direction.}
    \label{fig: tmp2}
\end{figure}

\subsubsection{AE+SINDy architecture}
\noindent The AE architecture is such that the encoder block consists of 4 hidden layers of widths $64, 32, 16$ and $8$, respectively, while the decoder has a symmetrical structure. 
Given a low Reynolds number, the dynamical system evolves onto a two-dimensional parabolic manifold \cite{loiseau2021pod, noack2003hierarchy}. To further account for the parametric dependency of the manifold on the Reynolds number, the number of latent variables, which correspond to the dimension of the bottleneck of the AE, is set to 3.
Regarding the SINDy algorithm, we employ a set of polynomials features $\vb{\Theta}$, up to the third degree, with respect to both the latent variables $\vb{z} = \left[z_1, z_2, z_3 \right]^T$ and the parameter $\frac{1}{Re}$ to approximate the latent dynamics $\vb{f}$, such that
\begin{equation}
    \dot{\vb{z}} = \vb{f}\left(\vb{z},\frac{1}{Re}\right) \approx \vb{\Theta}\left(\vb{z},\frac{1}{Re}\right)\vb{\Xi},
    \label{eq: latent_eq_fluid}
\end{equation}
where $\mathbf{\Xi}$ is the sparse matrix of the unknown multiplicative coefficients of the features of library $\mathbf{\Theta}$.
The parametric dependence is with respect to the reciprocal of Reynolds number -- which corresponds to the viscosity $\nu$ -- for consistency with the full order equation \eqref{eq: NS}.
\paolo{Differently from the previous example in  Sect.~\ref{sec:beam} for which the parametric dependency is directly related to the forcing term, thus known a priori, this application provides an example for which no information about how the parameter enters the latent equations is available. For this reason, we consider a extensive polynomial library that includes also parameter interactions with latent variables.\\
The identified coefficients of the latent system \eqref{eq: latent_eq_fluid}, i.e. the entries of $\vb{\Xi}$, are reported in Table \ref{tab:coeff_fluid}.
}

\begin{table}[b!]
\setlength{\tabcolsep}{4.5pt}
\renewcommand{\arraystretch}{1.05}
\caption{\paolo{Entries of $\vb{\Xi}$ obtained by training AE+SINDy. Each column indicates to which candidate feature 
the coefficient is associated to, while rows indicate the left-hand-side of the equation of the system \eqref{eq: latent_eq_fluid} in which the coefficient enters. Note that the following scaling: $\beta=10^3/Re  =\nu \cdot 10^3$ has been considered.}}
\begin{tabular}{lcccccccccccc}
    & $1$ & $z_1$ & $z_2$ & $z_3$ & $\beta$ & $z_1^2$ & $z_1z_2$ & $z_1z_3$ & $z_1\beta$ & $z_2^2$ & $z_2z_3$ & $z_2\beta$ \\ \hline
\multicolumn{1}{l|}{$\dot{z}_1$}& 2.23      & 0       & 3.44      & 61.8       & -0.25       & 0        & -4.17       & 0       & 0      & 0       & -8.15       & -0.43      \\
\multicolumn{1}{l|}{$\dot{z}_2$} & 10.81       & -89.0       & 112.5      & -67.5       & 0       & 327.2        & -623.2       & 82.17       & -2.68      & 244.7       & -54.0       & -1.49       \\
\multicolumn{1}{l|}{$\dot{z}_3$}& 0       & -35.4      & -16.1      & 0       & 0.04       & 229.2        & -61.2       & 81.9       & 2.34     & -7.41      & 1.46       & 1.77       \\
\vspace{-5pt}
\\

   & $z_3^2$ & $z_3\beta$ & $\beta^2$ & $z_1^3$ & $z_1^2z_2$ & $z_1^2z_3$ & $z_1^2\beta$ & $z_1z_2^2$ & $z_1z_2z_3$ & $z_1z_2\beta$ & $z_1z_3^2$  & $z_1z_3\beta$  \\ \hline
\multicolumn{1}{l|}{$\dot{z}_1$}&  -7.55  & -29.5      & -8.15       & 0      & -33.6        & 111.3       & -1175.1      & 0.14      & -142.5      &  1449.0     & -3.38       & -687.0       \\
\multicolumn{1}{l|}{$\dot{z}_2$}&  -3.79 & -23.0       & 3.96       & 0      & -734.1        & 1309.0      & -187.5        & 12.5       & -781.0       & 0      & -2.23      & 0         \\
\multicolumn{1}{l|}{$\dot{z}_3$}& 31.0  & 11.8       & 0       & 0      & 0        & 154.1       & 217.0        & -12.4       & -153.8      & -10.9     & 0       & 0         \\
\vspace{-5pt}
\\
   & $z_1\beta^2$ & $z_2^3$ & $z_2^2z_3$ & $z_2^2\beta$ & $z_2z_3^2$ & $z_2z_3\beta$ & $z_2\beta^2$ & $z_3^3$ & $z_3^2\beta$ & $z_3\beta^2$  & $\beta^3$ &\\ \cline{1-12} 
\multicolumn{1}{l|}{$\dot{z}_1$}  &0       & 4.58       & -565.0      & 0        & 408.2       & -18.5        & 0       & -262.1       & 7.91      & 0       & 0        &   \\
\multicolumn{1}{l|}{$\dot{z}_2$}& 0       & 151.5       & 113.0      & -2.51        & -85.9      & 6.78        & 0       & 47.3       & 1.26      & 0   & 0          & \\
\multicolumn{1}{l|}{$\dot{z}_3$}&0       & 94.4       & -121.0      & 4.03        & 139.7       & -2.4        & 0       & -58.5      & 0.77      & 0    & 0  &

\end{tabular}
\label{tab:coeff_fluid}
\end{table}

\subsubsection{Results}
\paragraph*{Time-marching solutions}
\noindent The AE+SINDy model is trained on $\tilde{\vb{X}}_\text{train}, \dot{\tilde{\vb{X}}}_\text{train}$ data, with the method introduced in Sect.~\ref{sect: offline_training}, then it is evaluated on the testing values $Re = \{43,55\}$ following the strategy in Sect.~\ref{sect: online_testing}. 
Approximated solutions are predicted by AE+SINDy up to final time $T = 30s$ and compared to the numerical values ${\vb{X_\text{test}}}$ in Fig.~\ref{fig: tmp2}. The proposed method provides solutions of comparable accuracy to the numerical one, though, at extremely advantageous computational cost and time. 
Furthermore, we emphasize how indeed the model manages to accurately reconstruct the different behaviors of the fluid, i.e.\ laminar and unsteady, as the Reynolds number varies.

However, this is not sufficient to ensure that the identified system is indeed representative of the whole dynamics of the phenomenon. To this end, we pass the identified system to the continuation algorithms to estimate the latent dynamic behavior of the system as the Reynolds number changes.

\paragraph*{Continuation}
\noindent We now intend to apply the continuation algorithm to the identified latent system in order to gain insight into the dynamics of the system under observation. Such approach allows us to directly compute the steady-state response, i.e.\ when the flow is fully developed, avoiding the costs of computing the transient and the error accumulation of numerical integration. In addition, the continuation technique is an extremely useful tool in this context for identifying and characterizing the transition of the dynamics from laminar to unsteady. Indeed, while in the unsteady regime the steady-state solution is characterized by periodic oscillations, in the laminar regime it is stationary. Therefore, we expect the continuation technique, which assumes a-priori the existence of a periodic response, to break down during a downward sweep over the Reynolds 
number when reaching the bifurcation point (see Figure~\ref{fig: continuation_rec}). We aim to verify whether we can effectively recover the expected bifurcation value, which is between $Re = 49$ and $50$ \cite{zdravkovich1997flow, rajani2009numerical}, and to check if the latent solutions estimated by continuation are physically representative of the dynamics of the phenomenon when projected back to the physical space. We stress that appropriately locating the bifurcation value would not be straightforward with time-marching schemes, since they would require instead the calculation of the entire time evolution of solutions for a dense sampling of parameter instances.

\begin{figure}[b!]
	\centering
	{\includegraphics[width=0.65\textwidth]{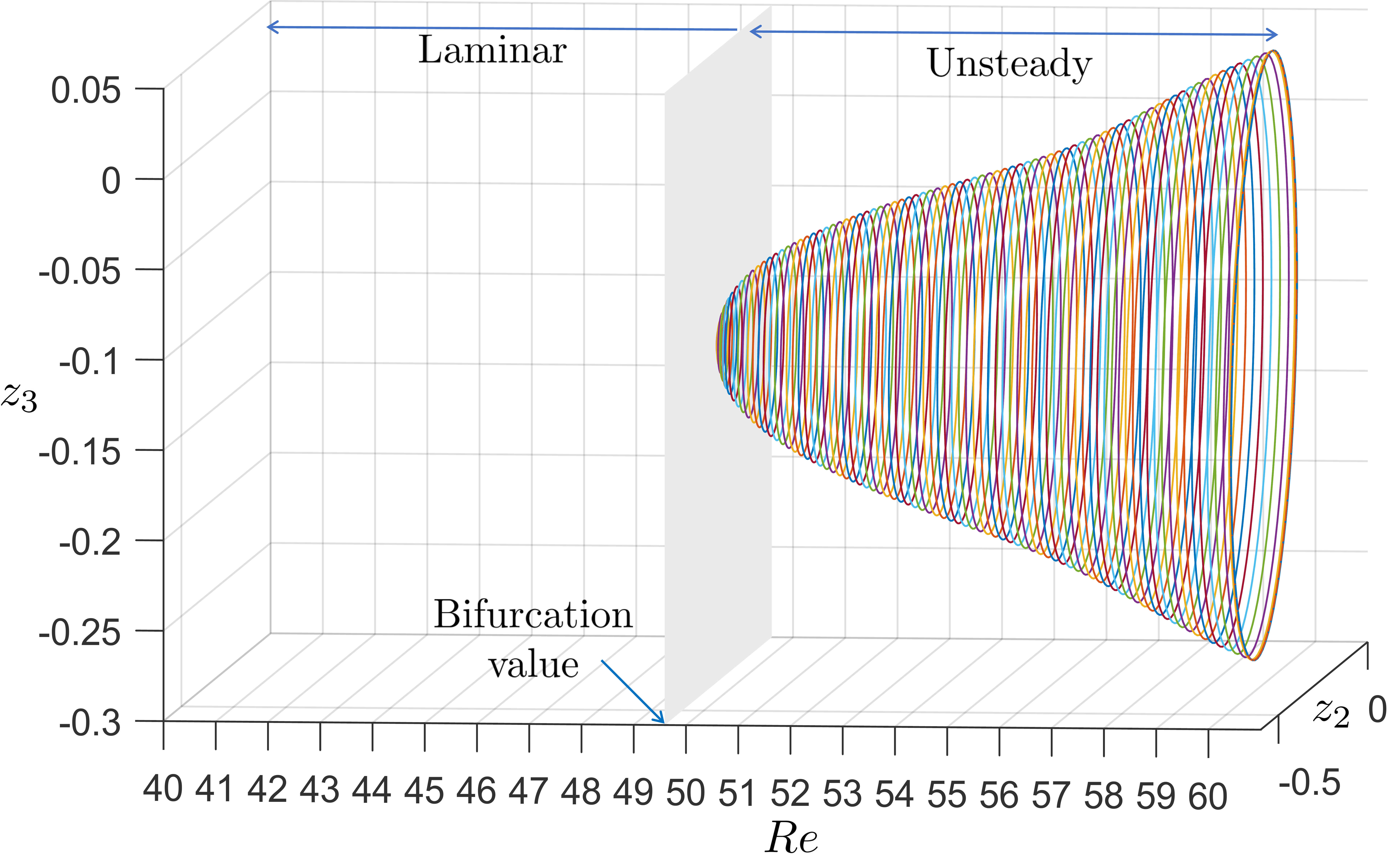}}
	\caption{Behaviour of the latent coordinates $(z_2,z_3)$ with respect to the $Re$ parameter. After a critical $Re$ value the fixed point solution (steady state flow) become unstable and a limit cycle arises (Hopf bifurcation). The limit cycle is embedded by the SINDy model integrated with continuation approach.}
    \label{fig: orbits_latent}
\end{figure}

\begin{figure}[p]
    \centering
    {\includegraphics[width=0.95\textwidth]{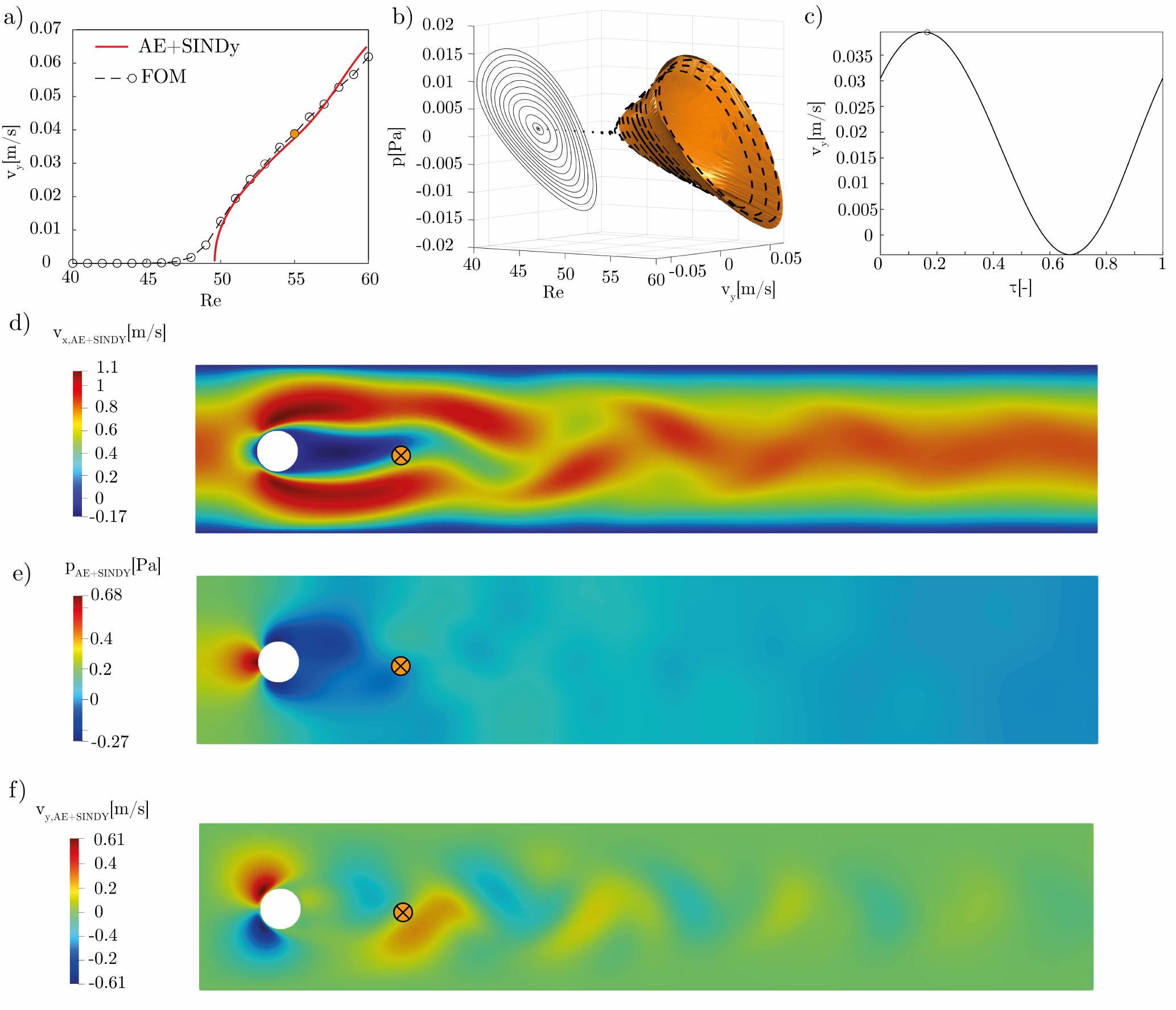}}
    \caption{
    Continuation solutions with AE+SINDy. Fig.~a) bifurcation diagram obtained from continuation compared with the FOM solution close to the steady state, the orange marker  highlights the point having \paolo{$Re=55$} used as testing parameter value
    Fig.~b) Surface described by the orbits envelope in the space ($Re,v_y,P$) referring to the node (highlighted previously) compared with the FOM solutions. Since the FOM is based on time-marching methods, the orbits here represented with dashed lines correspond to the one closer to the steady state. To better highlight the limit cycles, the orbits are also projected in the plane ($v_y,P$).
    Fig.~c) solution corresponding to \paolo{$Re=55$} achieved with AE+SINDy, the marker highlights the time instant kept as reference in the spatial reconstructions
    Fig.~d)-e)-f) AE+SINDy velocity fields along the two directions and the pressure field with \paolo{$Re=55$}. The marker highlights the mesh node used as reference.}
    \label{fig: continuation_rec}
\end{figure}

Starting from $Re=60$, we employ the continuation algorithm to compute the periodic steady-state latent solutions over the parametric domain under consideration, namely for $Re \in \mathcal{P} = [40,60]$. We obtain that continuation procedure correctly halts at the Reynolds number which matches the expected bifurcation value for the transition from unsteady to laminar regime (see Fig.~\ref{fig: orbits_latent}). To convey continuation insights to the original physical system we follow the procedure presented in Sect.~\ref{sect: continuation}: physical steady-state solutions are reconstructed by mapping the so-computed periodic latent variables via decoder and then by projection to the physical space through POD modes. The comparison between the predicted physical orbits with the ones drawn from the original data is illustrated in Fig.~\ref{fig: continuation_rec}.

Moreover, this technique allows to draw bifurcation diagrams which describes the characteristics of the flow as the Reynolds number changes. For instance, we report in Fig.~\ref{fig: continuation_rec}a, the maximum value over time of the velocity of the flow in the cross-channel direction for a spatial point, as a function of the parameter $Re$. This velocity component is a direct indicator of the presence of vortexes in the flow that characterize the unsteady regime, while it is absent for the laminar flow.
We observe that the bifurcation diagram identified by continuation is coherent with the (training and testing) snapshot data of the physical system (see Fig.~\ref{fig: continuation_rec}a).
For completeness, moreover, we show in Fig.~\ref{fig: continuation_rec}d-e-f the physical reconstruction of the steady-state solution obtained by continuation at \paolo{$Re = 55$}. Comparing it with the numerical solution (see Fig.~\ref{fig: tmp2}i-j-k), we observe that the dynamic behavior of the system is accurately detected.

\section{Conclusions}
\label{sec:conclusions}
\noindent In this work we proposed a framework for building low-dimensional dynamical models which overcomes the major bottlenecks of full order models to approximate solutions of parametrized, nonlinear, time-dependent systems of PDEs. Our method exploits both POD and AEs to construct a ROM, while simultaneously identifying the reduced dynamics via parametric SINDy.
The possibility to incorporate parametric dependences in the latent ODE system, besides extending existing approaches \cite{champion2019data, bakarji2022discovering}, opens the door to the application of continuation techniques. On one hand, these techniques allow to derive a portrait of the dynamics of the observed system; on the other hand, they offer the chance to efficiently estimate periodic steady-state solutions, without computing the whole transient dynamics.

We applied the proposed method to study two problems, the first in structural mechanics dealing with the hardening behaviour of a clamped clamped beam, and the second in fluid dynamics dealing with the motion of a fluid flow past a cylinder. In the first application, starting from simulation data covering a limited portion of the transient dynamics, we could efficiently derive a one-dimensional dynamical ROM which predicts the steady-state responses for new parameters outside the training range, thus extrapolating in time for long-term horizons while simultaneously generalizing with respect to system parameters. For the second example, the identified latent model allows to accurately estimate full order solutions, both for laminar and unsteady behaviours, as well as to identify the correct value of the Reynolds number at which the change of the flow regime occurs.

\paolo{The considered examples highlight that if AE+SINDy successfully converges to a solution on training instances, it also allows numerical integration in long time horizon for testing instances as well as continuation algorithms to work, hence implying robustness and stability with respect to temporal and parameter extrapolation. This suggests that the AE+SINDy algorithm is less prone to overfitting phenomena with respect to standard data-driven method, thanks to the incorporation of physical and dynamical information via SINDy.} 

In addition to ensuring high accuracy at low computational cost, this method is highly non-intrusive since it just requires a very limited number of snapshots solutions and a choice of basis functions for the latent dynamics, instead of accessing the expensive FOM operators. Therefore, even though only simulation data are considered in the present work, the proposed method can be directly applied to observational and experimental data sets as well. This would allow to construct ROMs and discover the underlying dynamics also in the case of systems for which no information about their generative model is available.

\paolo{Further improvements of the proposed method include the use of more powerful neural networks than standard feedforward autoencoders and the implementation of more sophisticated techniques for sparsity imposition in order to promote parsimonia and interpretability of the reduced dynamical model.}


\section*{Acknowledgment}
\noindent
Paolo Conti has been supported on funding under the JRC STEAM STM-Politecnico di Milano agreement. Stefania Fresca, Giorgio Gobat, and Andrea Manzoni have been supported by Fondazione Cariplo, Grant no. 2019-4608. 
The authors would like to express their appreciation to Dr. Andrea Opreni for the fruitful discussions.









\bibliographystyle{abbrv}
\bibliography{main.bib}

\begin{thebibliography}{10}

\bibitem{amsallem2012nonlinear}
D.~Amsallem, M.~J. Zahr, and C.~Farhat.
\newblock Nonlinear model order reduction based on local reduced-order bases.
\newblock {\em International Journal for Numerical Methods in Engineering},
  92(10):891--916, 2012.

\bibitem{ananthkrishnan1996characterization}
N.~Ananthkrishnan and K.~Sudhakar.
\newblock Characterization of periodic motions in aircraft lateral dynamics.
\newblock {\em Journal of guidance, control, and dynamics}, 19(3):680--685,
  1996.

\bibitem{bakarji2022discovering}
J.~Bakarji, K.~Champion, J.~N. Kutz, and S.~L. Brunton.
\newblock Discovering governing equations from partial measurements with deep
  delay autoencoders.
\newblock {\em arXiv preprint arXiv:2201.05136}, 2022.

\bibitem{benner2017model}
P.~Benner, M.~Ohlberger, A.~Cohen, and K.~(Eds.)~Willcox.
\newblock {\em Model reduction and approximation: theory and algorithms}.
\newblock SIAM, 2017.

\bibitem{brunton2016discovering}
S.~L. Brunton, J.~L. Proctor, and J.~N. Kutz.
\newblock Discovering governing equations from data by sparse identification of
  nonlinear dynamical systems.
\newblock {\em Proceedings of the national academy of sciences},
  113(15):3932--3937, 2016.

\bibitem{brunton2016sparse}
S.~L. Brunton, J.~L. Proctor, and J.~N. Kutz.
\newblock Sparse identification of nonlinear dynamics with control (sindyc).
\newblock {\em IFAC-PapersOnLine}, 49(18):710--715, 2016.

\bibitem{callaham2022role}
J.~L. Callaham, S.~L. Brunton, and J.-C. Loiseau.
\newblock On the role of nonlinear correlations in reduced-order modelling.
\newblock {\em Journal of Fluid Mechanics}, 938, 2022.

\bibitem{champion2019data}
K.~Champion, B.~Lusch, J.~N. Kutz, and S.~L. Brunton.
\newblock Data-driven discovery of coordinates and governing equations.
\newblock {\em Proceedings of the National Academy of Sciences},
  116(45):22445--22451, 2019.

\bibitem{conti2022multi}
P.~Conti, M.~Guo, A.~Manzoni, and J.~S. Hesthaven.
\newblock Multi-fidelity surrogate modeling using long short-term memory
  networks.
\newblock {\em Computer methods in applied mechanics and engineering},
  404:115811, 2023.

\bibitem{corigliano2004mechanical}
A.~Corigliano, B.~De~Masi, A.~Frangi, C.~Comi, A.~Villa, and M.~Marchi.
\newblock Mechanical characterization of polysilicon through on-chip tensile
  tests.
\newblock {\em Journal of Microelectromechanical Systems}, 13(2):200--219,
  2004.

\bibitem{cybenko1989approximation}
G.~Cybenko.
\newblock Approximation by superpositions of a sigmoidal function.
\newblock {\em Mathematics of control, signals and systems}, 2(4):303--314,
  1989.

\bibitem{dankowicz2013recipes}
H.~Dankowicz and F.~Schilder.
\newblock {\em Recipes for continuation}.
\newblock SIAM, 2013.

\bibitem{detroux2015harmonic}
T.~Detroux, L.~Renson, L.~Masset, and G.~Kerschen.
\newblock The harmonic balance method for bifurcation analysis of large-scale
  nonlinear mechanical systems.
\newblock {\em Computer Methods in Applied Mechanics and Engineering},
  296:18--38, 2015.

\bibitem{dhooge2006matcont}
A.~Dhooge, W.~Govaerts, Y.~A. Kuznetsov, W.~Mestrom, A.~Riet, and B.~Sautois.
\newblock Matcont and cl matcont: Continuation toolboxes in matlab.
\newblock {\em Universiteit Gent, Belgium and Utrecht University, The
  Netherlands}, 2006.

\bibitem{doedel2007auto}
E.~J. Doedel, A.~R. Champneys, F.~Dercole, T.~F. Fairgrieve, Y.~A. Kuznetsov,
  B.~Oldeman, R.~Paffenroth, B.~Sandstede, X.~Wang, and C.~Zhang.
\newblock Auto-07p: Continuation and bifurcation software for ordinary
  differential equations.
\newblock 2007.

\bibitem{epstein1996nonlinear}
I.~R. Epstein and K.~Showalter.
\newblock Nonlinear chemical dynamics: oscillations, patterns, and chaos.
\newblock {\em The Journal of Physical Chemistry}, 100(31):13132--13147, 1996.

\bibitem{franco2021deep}
N.~Franco, A.~Manzoni, and P.~Zunino.
\newblock A deep learning approach to reduced order modelling of parameter
  dependent partial differential equations.
\newblock {\em Mathematics of Computation}, 92(340):483--524, 2023.

\bibitem{fresca2021comprehensive}
S.~Fresca, L.~Dede, and A.~Manzoni.
\newblock A comprehensive deep learning-based approach to reduced order
  modeling of nonlinear time-dependent parametrized pdes.
\newblock {\em Journal of Scientific Computing}, 87(2):1--36, 2021.

\bibitem{fresca2022deep}
S.~Fresca, G.~Gobat, P.~Fedeli, A.~Frangi, and A.~Manzoni.
\newblock Deep learning-based reduced order models for the real-time simulation
  of the nonlinear dynamics of microstructures.
\newblock {\em International Journal for Numerical Methods in Engineering},
  123(20):4749--4777, 2022.

\bibitem{fresca2021real}
S.~Fresca and A.~Manzoni.
\newblock Real-time simulation of parameter-dependent fluid flows through deep
  learning-based reduced order models.
\newblock {\em Fluids}, 6(7):259, 2021.

\bibitem{fresca2022pod}
S.~Fresca and A.~Manzoni.
\newblock Pod-dl-rom: enhancing deep learning-based reduced order models for
  nonlinear parametrized pdes by proper orthogonal decomposition.
\newblock {\em Computer Methods in Applied Mechanics and Engineering},
  388:114181, 2022.

\bibitem{fries2022lasdi}
W.~D. Fries, X.~He, and Y.~Choi.
\newblock Lasdi: Parametric latent space dynamics identification.
\newblock {\em Computer Methods in Applied Mechanics and Engineering},
  399:115436, 2022.

\bibitem{fukami2021sparse}
K.~Fukami, T.~Murata, K.~Zhang, and K.~Fukagata.
\newblock Sparse identification of nonlinear dynamics with low-dimensionalized
  flow representations.
\newblock {\em Journal of Fluid Mechanics}, 926, 2021.

\bibitem{gobat2022virtual}
G.~Gobat, S.~Fresca, A.~Manzoni, and A.~Frangi.
\newblock Virtual twins of nonlinear vibrating multiphysics microstructures:
  physics-based versus deep learning-based approaches.
\newblock {\em arXiv preprint arXiv:2205.05928}, 2022.

\bibitem{gobat2022reduced}
G.~Gobat, A.~Opreni, S.~Fresca, A.~Manzoni, and A.~Frangi.
\newblock Reduced order modeling of nonlinear microstructures through proper
  orthogonal decomposition.
\newblock {\em Mechanical Systems and Signal Processing}, 171:108864, 2022.

\bibitem{gonzalez2018deep}
F.~J. Gonzalez and M.~Balajewicz.
\newblock Deep convolutional recurrent autoencoders for learning
  low-dimensional feature dynamics of fluid systems.
\newblock {\em arXiv preprint arXiv:1808.01346}, 2018.

\bibitem{goodfellow2016deep}
I.~Goodfellow, Y.~Bengio, and A.~Courville.
\newblock {\em Deep learning}.
\newblock MIT press, 2016.

\bibitem{goyal2021learning}
P.~Goyal and P.~Benner.
\newblock Learning low-dimensional quadratic-embeddings of high-fidelity
  nonlinear dynamics using deep learning.
\newblock {\em arXiv preprint arXiv:2111.12995}, 2021.

\bibitem{goyal2022discovery}
P.~Goyal and P.~Benner.
\newblock Discovery of nonlinear dynamical systems using a runge--kutta
  inspired dictionary-based sparse regression approach.
\newblock {\em Proceedings of the Royal Society A}, 478(2262):20210883, 2022.

\bibitem{guillot2019taylor}
L.~Guillot, B.~Cochelin, and C.~Vergez.
\newblock A taylor series-based continuation method for solutions of dynamical
  systems.
\newblock {\em Nonlinear dynamics}, 98(4):2827--2845, 2019.

\bibitem{guillot2020purely}
L.~Guillot, A.~Lazarus, O.~Thomas, C.~Vergez, and B.~Cochelin.
\newblock A purely frequency based floquet-hill formulation for the efficient
  stability computation of periodic solutions of ordinary differential systems.
\newblock {\em Journal of Computational Physics}, 416:109477, 2020.

\bibitem{guillot2017continuation}
L.~Guillot, P.~Vigu{\'e}, C.~Vergez, and B.~Cochelin.
\newblock Continuation of quasi-periodic solutions with two-frequency harmonic
  balance method.
\newblock {\em Journal of Sound and Vibration}, 394:434--450, 2017.

\bibitem{HesthavenRozzaStamm}
J.~Hesthaven, G.~Rozza, and B.~Stamm.
\newblock {\em Certified Reduced Basis Methods for Parametrized Partial
  Differential Equations}.
\newblock SpringerBriefs in Mathematics. Springer, 2016.

\bibitem{hornik1989multilayer}
K.~Hornik, M.~Stinchcombe, and H.~White.
\newblock Multilayer feedforward networks are universal approximators.
\newblock {\em Neural networks}, 2(5):359--366, 1989.

\bibitem{kaiser2018sparse}
E.~Kaiser, J.~N. Kutz, and S.~L. Brunton.
\newblock Sparse identification of nonlinear dynamics for model predictive
  control in the low-data limit.
\newblock {\em Proceedings of the Royal Society A}, 474(2219):20180335, 2018.

\bibitem{kalia2021learning}
M.~Kalia, S.~L. Brunton, H.~G. Meijer, C.~Brune, and J.~N. Kutz.
\newblock Learning normal form autoencoders for data-driven discovery of
  universal, parameter-dependent governing equations.
\newblock {\em arXiv preprint arXiv:2106.05102}, 2021.

\bibitem{kim2022fast}
Y.~Kim, Y.~Choi, D.~Widemann, and T.~Zohdi.
\newblock A fast and accurate physics-informed neural network reduced order
  model with shallow masked autoencoder.
\newblock {\em J. Comput. Phys.}, 451:110841, 2022.

\bibitem{kingma2014adam}
D.~P. Kingma and J.~Ba.
\newblock Adam: A method for stochastic optimization.
\newblock {\em arXiv preprint arXiv:1412.6980}, 2014.

\bibitem{kneifl2021nonintrusive}
J.~Kneifl, D.~Grunert, and J.~Fehr.
\newblock A nonintrusive nonlinear model reduction method for structural
  dynamical problems based on machine learning.
\newblock {\em International Journal for Numerical Methods in Engineering},
  122(17):4774--4786, 2021.

\bibitem{krack2019harmonic}
M.~Krack and J.~Gross.
\newblock {\em Harmonic balance for nonlinear vibration problems}, volume~1.
\newblock Springer, 2019.

\bibitem{krauskopf2007numerical}
B.~Krauskopf, H.~M. Osinga, and J.~Gal{\'a}n-Vioque.
\newblock {\em Numerical continuation methods for dynamical systems}, volume~2.
\newblock Springer, 2007.

\bibitem{PyCont}
D.~LaMar.
\newblock Pycont.
\newblock {\em https://pydstool.github.io/PyDSTool/PyCont}, 2021.

\bibitem{lee2020model}
K.~Lee and K.~T. Carlberg.
\newblock Model reduction of dynamical systems on nonlinear manifolds using
  deep convolutional autoencoders.
\newblock {\em J. Comput. Phys.}, 404:108973, 2020.

\bibitem{leshno1993multilayer}
M.~Leshno, V.~Y. Lin, A.~Pinkus, and S.~Schocken.
\newblock Multilayer feedforward networks with a nonpolynomial activation
  function can approximate any function.
\newblock {\em Neural networks}, 6(6):861--867, 1993.

\bibitem{loiseau2021pod}
J.-C. Loiseau, S.~Brunton, and B.~Noack.
\newblock From the pod-galerkin method to sparse manifold models, 2021.

\bibitem{lusch2018deep}
B.~Lusch, J.~N. Kutz, and S.~L. Brunton.
\newblock Deep learning for universal linear embeddings of nonlinear dynamics.
\newblock {\em Nature communications}, 9(1):1--10, 2018.

\bibitem{malvern1969introduction}
L.~E. Malvern.
\newblock {\em Introduction to the Mechanics of a Continuous Medium}.
\newblock Number Monograph. 1969.

\bibitem{maulik2021reduced}
R.~Maulik, B.~Lusch, and P.~Balaprakash.
\newblock Reduced-order modeling of advection-dominated systems with recurrent
  neural networks and convolutional autoencoders.
\newblock {\em Phys. Fluids}, 33(3):037106, 2021.

\bibitem{negri2016redbkit}
F.~Negri.
\newblock redbkit version 2.2.
\newblock {\em http://redbkit.github.io/redbKIT}, 2016.

\bibitem{noack2003hierarchy}
B.~R. Noack, K.~Afanasiev, M.~MORZY{\'N}SKI, G.~Tadmor, and F.~Thiele.
\newblock A hierarchy of low-dimensional models for the transient and
  post-transient cylinder wake.
\newblock {\em Journal of Fluid Mechanics}, 497:335--363, 2003.

\bibitem{opreni2021analysis}
A.~Opreni, N.~Boni, R.~Carminati, and A.~Frangi.
\newblock Analysis of the nonlinear response of piezo-micromirrors with the
  harmonic balance method.
\newblock {\em Actuators}, 10(2):21, 2021.

\bibitem{opreni2021model}
A.~Opreni, A.~Vizzaccaro, A.~Frangi, and C.~Touz{\'e}.
\newblock Model order reduction based on direct normal form: application to
  large finite element mems structures featuring internal resonance.
\newblock {\em Nonlinear Dynamics}, 105(2):1237--1272, 2021.

\bibitem{opreni2022high}
A.~Opreni, A.~Vizzaccaro, C.~Touz{\'e}, and A.~Frangi.
\newblock High-order direct parametrisation of invariant manifolds for model
  order reduction of finite element structures: application to generic forcing
  terms and parametrically excited systems.
\newblock {\em Nonlinear Dynamics}, pages 1--47, 2022.

\bibitem{osborne1969shooting}
M.~R. Osborne.
\newblock On shooting methods for boundary value problems.
\newblock {\em Journal of mathematical analysis and applications},
  27(2):417--433, 1969.

\bibitem{pagani2018numerical}
S.~Pagani, A.~Manzoni, and A.~Quarteroni.
\newblock Numerical approximation of parametrized problems in cardiac
  electrophysiology by a local reduced basis method.
\newblock {\em Computer Methods in Applied Mechanics and Engineering},
  340:530--558, 2018.

\bibitem{quarteroni2015reduced}
A.~Quarteroni, A.~Manzoni, and F.~Negri.
\newblock {\em Reduced basis methods for partial differential equations: an
  introduction}, volume~92.
\newblock Springer, 2015.

\bibitem{rajani2009numerical}
B.~Rajani, A.~Kandasamy, and S.~Majumdar.
\newblock Numerical simulation of laminar flow past a circular cylinder.
\newblock {\em Applied Mathematical Modelling}, 33(3):1228--1247, 2009.

\bibitem{veltz2020bifurcationkit}
R.~Veltz.
\newblock {\em BifurcationKit. jl}.
\newblock PhD thesis, Inria Sophia-Antipolis, 2020.

\bibitem{vizzaccaro2022high}
A.~Vizzaccaro, A.~Opreni, L.~Salles, A.~Frangi, and C.~Touz{\'e}.
\newblock High order direct parametrisation of invariant manifolds for model
  order reduction of finite element structures: application to large amplitude
  vibrations and uncovering of a folding point.
\newblock {\em Nonlinear Dynamics}, 110(1):525--571, 2022.

\bibitem{vizzaccaro2021direct}
A.~Vizzaccaro, Y.~Shen, L.~Salles, J.~Blaho{\v{s}}, and C.~Touz{\'e}.
\newblock Direct computation of nonlinear mapping via normal form for
  reduced-order models of finite element nonlinear structures.
\newblock {\em Computer Methods in Applied Mechanics and Engineering},
  384:113957, 2021.

\bibitem{zdravkovich1997flow}
M.~M. Zdravkovich.
\newblock {\em Flow around circular cylinders: Volume 2: Applications},
  volume~2.
\newblock Oxford university press, 1997.

\bibitem{zega2020numerical}
V.~Zega, G.~Gattere, S.~Koppaka, A.~Alter, G.~D. Vukasin, A.~Frangi, and T.~W.
  Kenny.
\newblock Numerical modelling of non-linearities in mems resonators.
\newblock {\em Journal of Microelectromechanical Systems}, 29(6):1443--1454,
  2020.

\end{thebibliography}

\end{document}